\PassOptionsToPackage{table}{xcolor}
\documentclass[sigconf]{acmart}

\setcopyright{none}
\renewcommand\footnotetextcopyrightpermission[1]{}
\usepackage{subcaption}
\usepackage{booktabs} 
\usepackage{algorithm}
\usepackage{algorithmic}
\usepackage{multirow}
\usepackage{amsmath}
\usepackage{mathtools}
\usepackage{colortbl}
\usepackage{wrapfig}

\usepackage[capitalize,noabbrev]{cleveref}

\definecolor{lightpurple}{RGB}{230, 210, 250}

\begin{document}

\title{Beyond Language Priors: Diagnosing and Fixing Visual-Origin Hallucinations in Multimodal LLMs}

\author{Peiyang Xu}
\authornote{Co-First Authors, both authors contributed equally to this research.}
\email{peiyang2003@gmail.com}
\affiliation{%
  \institution{Tsinghua University}
  \city{Beijing}
  \country{China}
}

\author{Xiaopei Zhu}
\authornotemark[1]
\email{zxpthu@gmail.com}
\affiliation{%
  \institution{Tsinghua University}
  \city{Beijing}
  \country{China}
}

\author{Jun Zhu}
\email{dcszj@mail.tsinghua.edu.cn}
\affiliation{%
  \institution{Tsinghua University}
  \city{Beijing}
  \country{China}
}

\author{Xiaolin Hu}
\authornote{Corresponding author.}
\email{xlhu@tsinghua.edu.cn}
\affiliation{%
  \institution{Tsinghua University}
  \city{Beijing}
  \country{China}
}

\begin{abstract}
Existing research on object hallucination in multimodal large language models (MLLMs) predominantly attributes the problem to language priors such as over-reliance on textual co-occurrence statistics. We challenge this view by presenting quantitative evidence for a complementary, under-explored cause: \emph{visual-origin hallucination}, where hallucinations arise from incorrect visual feature extraction and misalignment between image and text embeddings. Through cosine similarity analysis and Smooth Grad-CAM entropy measurements, we show that hallucinated samples exhibit systematically lower image-text similarity (average 0.158 vs.\ $-$0.122) and inverted attention patterns, where attention is dispersed when the target object is present but wrongly concentrated when it is absent. Guided by this diagnosis, we propose Adversarial Contrastive Fine-Tuning (ACFT). ACFT uses an Adversarial Hallucination Attribute Flipping (AHAF) procedure, involving minimal, targeted adversarial perturbations that flip an image's hallucination attribute, to construct perfectly aligned positive-negative pairs, which are then used for contrastive fine-tuning. AHAF simultaneously serves as a \emph{diagnostic probe}, revealing that MLLM visual representations lie dangerously close to hallucination decision boundaries. Requiring only 0.9\% of the COCO dataset and adding zero inference overhead, ACFT achieves state-of-the-art performance on POPE, MME, and four description-level hallucination benchmarks across LLaVA, MiniGPT-4, and Qwen2.5-VL. Code is available at \url{https://github.com/zxp555/ACFT_MM26}.
\end{abstract}



\maketitle

\section{Introduction} \label{intro}
 
Object hallucination, where Multimodal Large Language Models (MLLMs) falsely perceive'' non-existent objects or ignore'' objects actually present in the image~\cite{rohrbach2018object,wei2024lehace}, remains a critical obstacle to deploying models such as LLaVA~\cite{liu2023llava}, MiniGPT-4~\cite{zhu2023minigpt4}, and GPT-4o~\cite{openai2023gpt4} in high-stakes domains. The dominant research narrative attributes hallucinations primarily to \emph{language priors}: over-reliance on textual co-occurrence~\cite{leng2023vcd,zhu2024ibd,chen2025perturbollavareducingmultimodalhallucinations}, hallucination heads attending to text tokens~\cite{zhou2024analyzingmitigatingobjecthallucination, yang2025mitigating}, summary-token attention sinks~\cite{huang2024opera}, or insufficient fine-grained reasoning supervision~\cite{zhang2024reflectiveinstructiontuningmitigating}. This language-prior perspective aligns naturally with long-output-text settings, where rich textual context amplifies linguistic biases.

We believe that this perspective is \emph{incomplete}. When model outputs are short, e.g., answering ``Is there an OBJ in the image?'', the model's reasoning may rely more on the visual modality, and a different hallucination mechanism emerges: one rooted in visual feature extraction itself~\cite{sun2025cause}. We term this \textbf{visual-origin hallucination} and provide a systematic evidence for its existence through two complementary analyses on LLaVA v1.5~\cite{liu2023llava} (see Figure~\ref{fig:cause} and SM for details):
 
\begin{figure*}[htbp]
\centering
\includegraphics[width=0.86\textwidth]{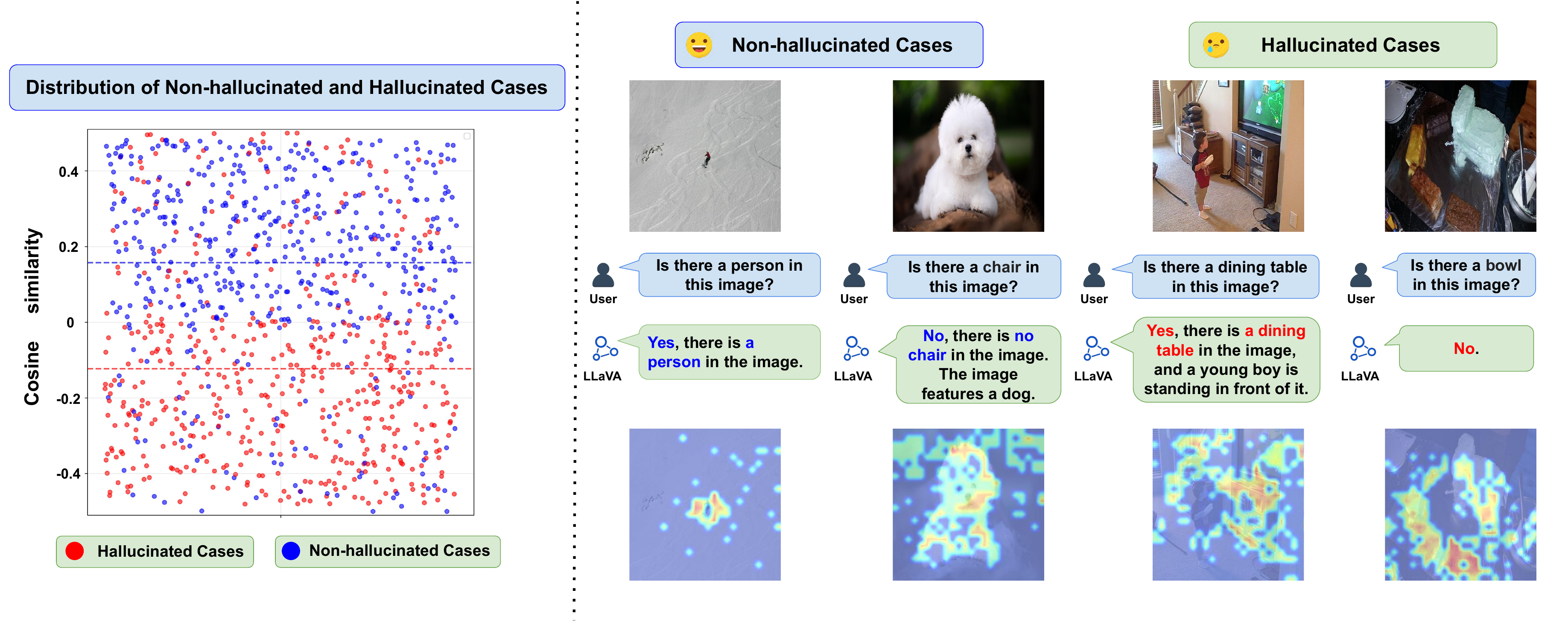}
\vspace{-6pt}
\Description{Cosine similarity distribution and Smooth Grad-CAM visualizations comparing hallucinated and non-hallucinated samples.}
\caption{Diagnosing visual-origin hallucination. \textbf{Left}: Cosine similarity distribution between image and text embeddings of non-hallucinated (\textcolor{blue}{blue dots}) and hallucinated (\textcolor{red}{red dots}) samples. The blue dashed line (\textcolor{blue}{- - -}) and red dashed line (\textcolor{red}{- - -}) show the average similarity.  \textbf{Right}: Case study using Smooth Grad-CAM. \textcolor{blue}{Blue} = correct responses, \textcolor{red}{red} = hallucinations. Hallucinated cases show inverted attention: dispersed when the object is present, concentrated when it is absent.}
\label{fig:cause}
\end{figure*}
 
\textit{Finding 1: Image--text embedding misalignment.} The cosine similarity between image and text embeddings is significantly lower for hallucinated samples than for correct ones (average 0.158 for correct cases vs.\ $-$0.122 for hallucinated cases), indicating a systematic breakdown in cross-modal alignment.
 
\textit{Finding 2: Inverted visual attention patterns.} We apply Smooth Grad-CAM~\cite{omeiza2019smoothgradcamenhancedinference,zhang2024redundancyrelevanceinformationflow} to visualize model attention and define a \emph{semantically appropriate} distribution: attention should focus on the target region when the object is present and disperse when it is absent. Hallucinated cases systematically violate this pattern (see Figure~\ref{fig:cause}, right). We quantify this via normalized Shannon entropy of each Grad-CAM heatmap over 500 hallucinated and 500 non-hallucinated samples: when the object is present, hallucinated models show 5.1\% higher entropy than correct ones (attention too dispersed, missing the target); when the object is absent, entropy is 6.2\% lower (wrongly concentrated on an irrelevant region, triggering hallucination).

\textit{Causal validation.} We further test causality through direct interventions on the visual encoder (see SM, Section 1.4). On POPE, Gaussian noise, downsampling, and a weaker encoder reduce average accuracy from 0.842 to 0.739--0.822, whereas a stronger SigLIP-SO400M encoder raises it to 0.864. These results provide causal support that visual feature quality drives object-existence hallucination.

These two findings establish that visual-origin hallucination is a distinct and measurable phenomenon linked to incorrect visual feature perception and image--text misalignment, particularly when textual outputs are short.
 
\textbf{From diagnosis to remedy.} Guided by these findings, we investigate contrastive learning to correct the identified visual misalignment. A na\"ive approach, ordinary contrastive fine-tuning (OCFT), pairs a matched image with a randomly selected unrelated image as the negative sample. However, OCFT yields unsatisfactory results (see Section~\ref{sec:OCFT}): the feature differences between positive and negative images are uncontrolled and not focused on the target object, making it difficult for the model to learn which visual features actually trigger hallucination.
 
To address this, we propose \textbf{Adversarial Contrastive Fine-Tuning (ACFT)}. The key idea is to use Adversarial Hallucination Attribute Flipping (AHAF), involving targeted adversarial perturbations via PGD that flip an image's hallucination attribute, to construct \emph{perfectly aligned} positive-negative pairs where the only difference is the controlled perturbation. This design has two advantages. First, because the pairs isolate exactly the visual features that trigger hallucination, the model can learn fine-grained rules for distinguishing hallucination-inducing from non-hallucinating inputs (we verify this via the target vs.\ non-target similarity gap in Figure~\ref{fig:ocft}). Second, AHAF doubles as a \emph{diagnostic probe}: the fact that minimal pixel-level perturbations suffice to flip the model's answer reveals that MLLM visual representations are alarmingly fragile, as they sit near hallucination decision boundaries even on clean images.

ACFT integrates directly into training with no inference overhead, unlike post-hoc methods~\cite{yin2024woodpecker,wu2024logic}, and requires only 0.9\% of the COCO dataset~\cite{lin2015microsoftcococommonobjects}---orders of magnitude less than full retraining~\cite{jiang2024hacl}. Experiments show that ACFT outperforms VCD~\cite{leng2023vcd}, OPERA~\cite{huang2024opera}, VTI~\cite{liu2024reducinghallucinationsvisionlanguagemodels}, and post-training methods (LLaVA-RLHF, DPO variants) on POPE~\cite{li2023evaluatingobjecthallucinationlarge} and MME~\cite{fu2024mmecomprehensiveevaluationbenchmark}, while preserving general visual comprehension.
 
Our contributions are: (1) We identify and provide a systematic evidence for visual-origin hallucination---a distinct hallucination mechanism driven by visual feature extraction errors and image--text misalignment, complementing the prevailing language-prior explanation. (2) We propose AHAF, which serves both as a diagnostic probe revealing the fragility of MLLM visual representations and as an efficient generator of aligned contrastive training data. (3) We propose ACFT, a data-efficient contrastive fine-tuning method that achieves state-of-the-art hallucination mitigation across multiple models and benchmarks with only 0.9\% of COCO data and zero inference overhead.

\section{Related Work}
 
\subsection{Multimodal Large Language Models (MLLMs)}
In recent years, Multimodal Large Language Models (MLLMs) have advanced rapidly. Proprietary models such as OpenAI's GPT-4o~\cite{openai2023gpt4} support both image and text inputs and show impressive performance across diverse multimodal tasks. Concurrently, numerous open-source MLLMs such as LLaVA~\cite{liu2023llava}, MiniGPT-4~\cite{zhu2023minigpt4}, etc.
have been introduced. Most of these models adopt a "visual encoder + large language model" architecture, achieving cross-modal alignment and reasoning capabilities through pretraining and instruction tuning.
 
\subsection{Mitigating Hallucinations for MLLMs}
 
Hallucination mitigation methods can be grouped by where they intervene. \emph{Input-level decoding} methods such as VCD~\cite{leng2023vcd} suppress hallucination-prone tokens by contrasting outputs from original versus perturbed images, and OPERA~\cite{huang2024opera} penalizes overconfident decoding patterns. \emph{Post-processing} methods like Woodpecker~\cite{yin2024woodpecker} verify and rewrite outputs using external grounding modules. \emph{Latent-space} methods such as VTI~\cite{liu2024reducinghallucinationsvisionlanguagemodels} steer hidden representations during inference. \emph{Post-training} methods include RLHF-based alignment~\cite{sun2023aligning} and DPO variants~\cite{yang2025opadpo,fu2025chip} that optimize preference over hallucinated vs.\ correct outputs.
 
A common thread across these approaches is their \emph{language-prior} perspective: they primarily address hallucinations arising from textual biases in long-output-text settings. Their effectiveness degrades in short-output-text scenarios where the model relies more on visual features. Our work complements this line by identifying visual-origin hallucination as a distinct mechanism and proposing a targeted fix via adversarial contrastive fine-tuning from the image-prior perspective.

\section{Methods}

\subsection{Mitigating Object Hallucination}
Our diagnosis in Section~\ref{intro} identifies two signatures of \emph{visual-origin hallucination}: image--text embedding misalignment and inverted attention patterns. We now describe a two-stage framework that directly targets both (Figure~\ref{fig:pipe}).
Stage one is AHAF, where we apply subtle adversarial perturbations on original images to construct aligned positive-negative image pairs. Stage two is ACFT, in which we design an adversarial contrastive loss function and fine-tune the MLLMs to mitigate object hallucination.

\subsection{Adversarial Hallucination Attribute Flipping}


To facilitate effective contrastive learning, we need to construct aligned positive–negative image pairs. In this study, we define a positive sample as an image that does not induce hallucinations in the model's output, and a negative sample as one that does. While one could manually collect natural images as positive and negative examples for contrast learning, doing so yields unsatisfactory results (See Section~\ref{sec:OCFT}). We believe the reason is that the feature differences between positive and negative images in OCFT are highly uncontrolled and not focused specifically on the target object itself, which makes it challenging for the MLLM to for attending to differences in the target object's features, thus impairing its ability to distinguish hallucination-inducing samples from non-hallucinating ones. To address this challenge, we propose an AHAF method to automatically construct aligned positive-negative image pairs and selectively alter key visual features for triggering targeted object hallucinations.

\textbf{AHAF as a diagnostic probe.} Beyond generating training data, AHAF reveals a striking property of MLLM visual representations: perturbations within a small $\ell_\infty$ ball suffice to flip the model's answer from correct to hallucinated. This indicates that the model's visual features for many objects lie near the hallucination decision boundary even on clean, unperturbed images---a finding that reinforces our diagnosis of visual-origin hallucination.



\begin{figure*}[htbp]
\centering
\includegraphics[width=0.86\textwidth]{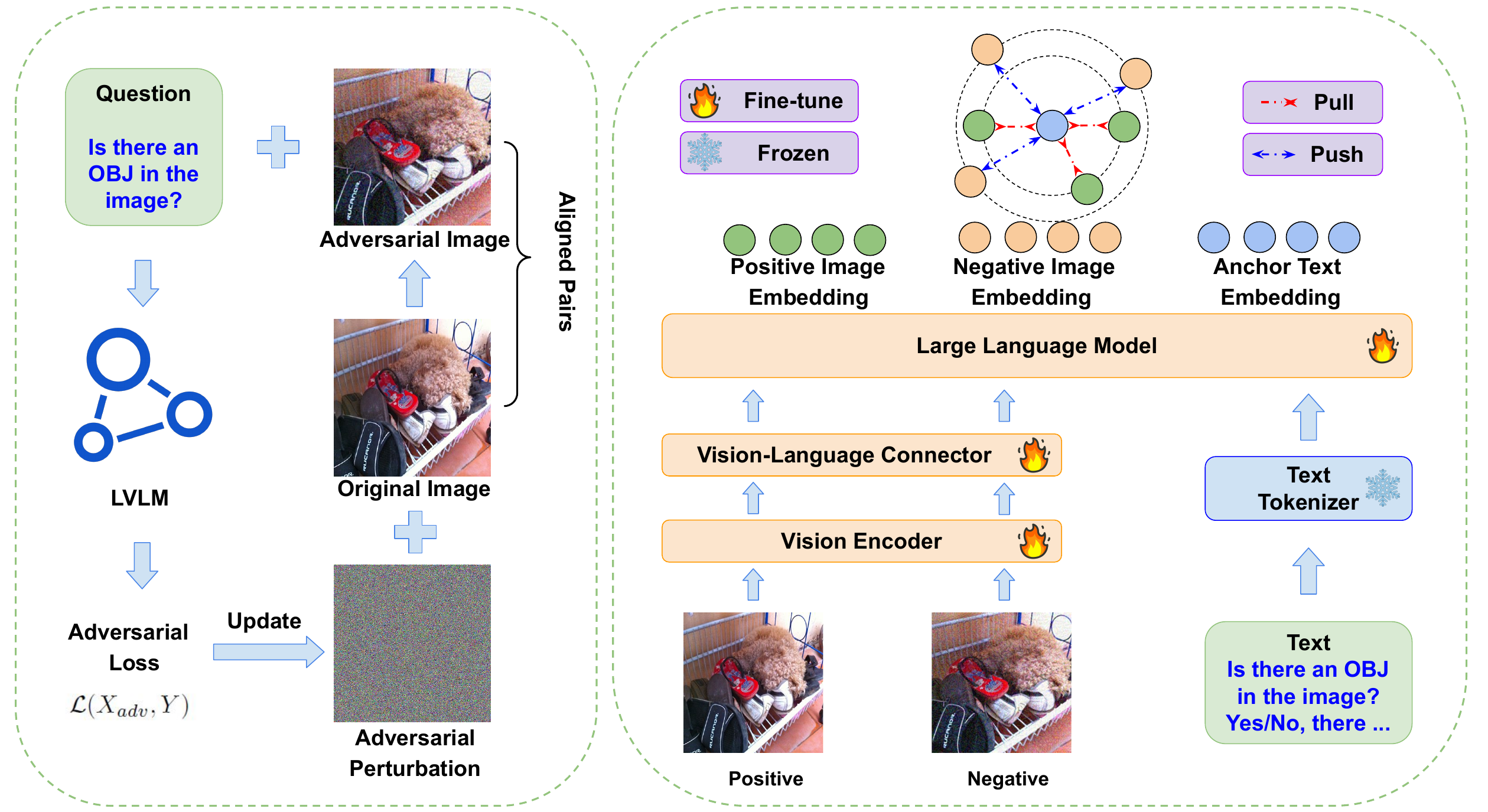}
\Description{Pipeline illustrating AHAF for generating adversarial positive-negative image pairs and ACFT for contrastive fine-tuning.}
\caption{Pipeline of our method. \textbf{Left}: AHAF applies minimal adversarial perturbations to original images, constructing aligned positive--negative pairs. The small perturbation budget required to flip the model's answer reveals the fragility of MLLM visual representations (diagnostic role). \textbf{Right}: ACFT maximizes the similarity between the text anchor and the positive image, while minimizing similarity with the negative image (remedial role).}
\label{fig:pipe}
\end{figure*}

As illustrated in Figure~\ref{fig:pipe}, AHAF first applies subtle adversarial perturbations generated by PGD~\cite{madry2019deeplearningmodelsresistant} method to the original images with its adversarial loss (Equation \ref{adv_loss}). These perturbations are then optimized to flip their hallucination attributes, converting a non-hallucinating image into one that induces object hallucination. The AHAF method thus automatically generates aligned positive–negative sample pairs in a highly targeted and efficient way. Because each pair differs only by the controlled adversarial perturbation, the resulting samples remain perfectly aligned, enabling precise, fine-grained analysis of the visual feature differences for the target object that gives rise to hallucination.

Next, we provide a detailed description of the AHAF method. Given an original image $X$ and a target text set $Y ={\{y_i\}}_{i=1}^{m}$, our objective is to generate an adversarial image $X_{adv}$ that, when input to the model, maximizes the model's likelihood of giving the target output. The optimization objective is formulated as:
\begin{equation}
X_{adv} := \mathop{\arg\min}\limits_{\hat{X}_{adv}\in \mathcal{B}} \sum_{i=1}^{m} -\log \left(p(y_i|\hat{X}_{adv})\right),
\end{equation}

To find suitable perturbations within the constrained space $\mathcal{B}$, PGD updates the image through the following iterative process:
{\small
\begin{equation}
\begin{aligned}
X^{t+1}_{adv} = \Pi_{\mathcal{B}(X,\epsilon)} \Big( & X^t_{adv} + \alpha \cdot \text{sign} \Big( \\
& \nabla_{X^t_{adv}} \sum_{i=1}^{m} -\log\big( p(y_i|X^t_{adv}) \big) \Big) \Big),
\end{aligned}
\end{equation}
}

In practice, we employ the cross-entropy loss to quantify the divergence between the model's output and the target text:
\begin{equation} \label{adv_loss}
\mathcal{L}(X_{adv}, Y) = -\sum_{i=1}^{m}\log\left(p(y_i|X_{adv})\right),
\end{equation}

After multiple iterations, the image $X_{adv}$ is used as a contrastive sample that can induce the model to produce an answer opposite to that of the original image.

\subsection{Adversarial Contrastive Fine-tuning}

Inspired by our preliminary findings (in Figure \ref{fig:cause}), we propose an ACFT method designed to enhance the alignment between visual and textual embedding and encourage the visual modality to focus on the correct image features, thus mitigating hallucinations. Based on the aligned positive–negative image pairs generated using AHAF method, the core of ACFT is to maximize the similarity between the text anchor and the positive image sample, while minimizing the similarity between the text anchor and the negative image sample in the embedding space through self-supervised contrast learning. The process of the ACFT algorithm is outlined in Algorithm~\ref{alg:acft}.

We design the ACFT method based on the following considerations. First, inspired by adversarial training~\cite{madry2018towards,bai2021recentadvancesadversarialtraining}, we inject adversarially optimized negative samples, which are tailored to exploit the target model's weaknesses, into the training process. The goal is to improve the model's ability to defend against strong perturbations and thereby learning more robust visual features that reduce hallucinations. Many studies \cite{li2024partimagenet++,li2023recognizing,liu2025comprehensive} have shown that adversarial training enhances model robustness to not only adversarial noise but also natural perturbations (e.g. illumination changes, blur, etc.).  Second, drawing on contrastive learning~\cite{radford2021clip, jiang2024hacl} principles, we construct aligned positive–negative sample pairs and apply a self supervised contrastive objective to sharpen the model's discrimination between hallucination inducing and non hallucinating images. Third, contrastive fine tuning offers strong generality, making ACFT applicable to a wide range of backbone architectures and ensuring both transferability and scalability. Finally, unlike full dataset retraining, our fine tuning approach requires only a modest amount of data, and unlike post post-processing technique, it imposes no extra inference-time overhead.

According to the above principles, we design an adversarial contrastive loss function $L_{\text{contra}}$ to measure similarity differences between positive and negative samples. To compute $L_{\text{contra}}$, we define a similarity-based contrastive objective between text and image representations. Specifically, for each anchor text $T_i$, we construct a corresponding positive-negative image pair $(X_i^+, X_i^-)$, where $X_i^+$ aligns with the text, while $X_i^-$ induces hallucination. Using the visual encoder $f(X)$ and text encoder $g(T)$, we extract image and text embeddings as $z^+ = f(X^+)$,  $z^- = f(X^-)$ and $t = g(T)$, respectively.
The adversarial contrastive loss for a single training instance is defined as:


{\small
\begin{equation}\label{eq:correct_contrast}
\begin{aligned}
\ell_{\text{contra}}^{(i)} = & -\log \big[\exp\left(\psi(t_i, z_i^+)/\tau\right)\big] + \log \\
& \Big[\exp\left(\psi(t_i, z_i^+)/\tau\right) + \exp\left(\psi(t_i, z_i^-)/\tau\right)\Big],
\end{aligned}
\end{equation}
}
where $\tau$ is a hyperparameter that controls the sharpness of the softmax distribution. The similarity between two vectors is defined as:
\begin{equation}\label{eq:sim}
\psi(\mathbf{x},\mathbf{y})
= \frac{\mathbf{x}^\top \mathbf{y}}{|\mathbf{x}| \cdot|\mathbf{y}|}.
\end{equation}

In batch training, where each batch contains $N$ text-image pairs, the batch-wise adversarial contrastive loss $L_{\text{contra}}$ is:

\begin{equation}\label{eq:batch_contrast}
L_{\text{contra}} = \frac{1}{N}\sum_{i=1}^{N}\ell_{\text{contra}}^{(i)},
\end{equation}

While mitigating hallucinations, we also aim to preserve the MLLM's original visual‐language generation capability. To this end, we introduce a classical cross‐entropy generation loss $L_{\text{gen}}$ during fine‐tuning. This loss measures the divergence between the model's predicted token distribution and the true distribution. Specifically, for a target sequence of length $T$, the model predicts the probability of the ground‐truth token $y_t^\ast$ conditioned on the preceding outputs $y_{<t}$  and the image features $I$ at each time step $t$. We then compute
\begin{equation}
\label{eq:crossentropy}
L_{\mathrm{gen}}
= -\frac{1}{T}\sum_{t=1}^{T}
\log p_\theta\bigl(y_t^\ast \mid y_{<t},\, X\bigr),
\end{equation}
which averages the negative log‐likelihood of the correct tokens over the entire sequence.

The overall objective is defined as:
\begin{equation}\label{eq:total_loss}
L_{\text{total}} \;=\; L_{\text{gen}} \;+\; \lambda \, L_{\text{contra}} \,,
\end{equation}
where $\lambda$ is a hyperparameter that is determined empirically.

\begin{algorithm}[htbp]
\small
\caption{Adversarial Contrast Fine-Tuning}
\label{alg:acft}
\textbf{Input}: Text inputs $\{T_i\}_{i=1}^N$, image pairs $\{(X_i^+, X_i^-)\}_{i=1}^N$\\
\textbf{Param}: $\tau$, $\lambda$, visual encoder $f(\cdot)$, text encoder $g(\cdot)$\\
\textbf{Output}: Fine-tuned model parameters\\
\begin{algorithmic}[1]
\FOR{each training iteration}
    \FOR{each triplet $(T_i, X_i^+, X_i^-)$ in batch}
        \STATE Extract embeddings: $t_i = g(T_i)$, $z_i^+ = f(X_i^+)$, $z_i^- = f(X_i^-)$.
        \STATE Compute similarities: $\mathrm{sim}(t_i, z_i^+)$ and $\mathrm{sim}(t_i, z_i^-)$.
        \STATE Compute contrastive loss $\ell_{\text{contra}}^{(i)}$ using Equation~\eqref{eq:correct_contrast}.
    \ENDFOR
    \STATE Compute batch contrastive loss using Equation~\eqref{eq:batch_contrast}.
    \FOR{each $(T_i, X_i^+)$ in batch}
        \STATE Perform generation task to compute $L_{\text{gen}}$ via Equation~\eqref{eq:crossentropy}.
    \ENDFOR
    \STATE Combine losses: $L_{\text{total}} = L_{\text{gen}} + \lambda L_{\text{contra}}$.
    \STATE Backpropagate and update model parameters using $L_{\text{total}}$.
\ENDFOR
\STATE \textbf{return} Fine-tuned model.
\end{algorithmic}
\end{algorithm}

\begin{table*}[htbp]
    \centering
    \resizebox{\textwidth}{!}{%
    \begin{tabular}{cc|cccc|cccc|cccc}
    \hline
    \multirow{2}{*}{\textbf{Subset}} & \multirow{2}{*}{\textbf{Method}} & \multicolumn{4}{c|}{\textbf{LlaVA v1.5 7B}} & \multicolumn{4}{c|}{\textbf{MiniGPT4 13B}} & \multicolumn{4}{c}{\textbf{Qwen2.5-VL 7B}} \\
    \cline{3-14}
    & & \textit{ACC} & \textit{Pre.} & \textit{Rec.} & \textit{F1} & \textit{ACC} & \textit{Pre.} & \textit{Rec.} & \textit{F1} & \textit{ACC} & \textit{Pre.} & \textit{Rec.} & \textit{F1} \\ \hline
    
    \multirow{6}{*}{{\textit{Adversarial}}}
    & origin & 0.779 & 0.721 & 0.911 & 0.805 & 0.700 & 0.670 & 0.791 & 0.725 & 0.864 & {0.940} & 0.778 & 0.851 \\
    & VCD & {0.808} & \textbf{0.847} & {0.753} & {0.797} & {0.734} & {0.701} & {0.817} & {0.754} & {0.868} & {0.925} & {0.800} & {0.858} \\
    & OPERA & {0.798} & {0.787} & {0.816} & {0.802} & {0.737} & {0.736} & {0.738} & {0.737} & {0.867} & \textbf{0.945} & {0.780} & {0.855} \\
    & Woodpecker & {0.771} & {0.710} & {\textbf{0.917}} & {0.800} & 0.741 & 0.678 & \textbf{0.917} & \textbf{0.780} & {0.854} & {0.903} & {0.793} & {0.845} \\
    & VTI & {0.805} & {0.770} & {0.871} & {0.817} & 0.700 & 0.668 & 0.795 & 0.726 & {0.866} & {0.942} & {0.780} & {0.853} \\
    & \cellcolor{lightpurple}Ours & \cellcolor{lightpurple}\textbf{0.841} & \cellcolor{lightpurple}0.802 & \cellcolor{lightpurple}0.905 & \cellcolor{lightpurple}\textbf{0.850} & \cellcolor{lightpurple}\textbf{0.771} & \cellcolor{lightpurple}\textbf{0.811} & \cellcolor{lightpurple}0.708 & \cellcolor{lightpurple}0.756 & \cellcolor{lightpurple}\textbf{0.877} & \cellcolor{lightpurple}0.897 & \cellcolor{lightpurple}\textbf{0.852} & \cellcolor{lightpurple}\textbf{0.874} \\ \hline
    
    \multirow{6}{*}{{\textit{Popular}}}
    & origin & 0.862 & 0.832 & 0.905 & 0.867 & 0.732 & 0.709 & 0.787 & 0.747 & 0.875 & {0.965} & 0.778 & 0.861 \\
    & VCD & {0.882} & \textbf{0.917} & {0.839} & {0.876} & {0.748} & {0.746} & {0.752} & {0.749} & {0.884} & {0.950} & {0.811} & {0.875} \\
    & OPERA & {0.886} & {0.847} & \textbf{0.940} & {0.891} & {0.737} & {0.715} & {0.789} & {0.750} & {0.881} & \textbf{0.973} & {0.783} & {0.868} \\
    & Woodpecker & {0.789} & {0.734} & {0.906} & {0.811} & 0.765 & 0.706 & \textbf{0.908} & 0.794 & {0.877} & {0.961} & {0.787} & {0.865} \\
    & VTI & {0.868} & {0.842} & {0.908} & {0.874} & 0.722 & 0.691 & 0.804 & 0.743 & {0.876} & {0.966} & {0.780} & {0.863} \\
    & \cellcolor{lightpurple}Ours & \cellcolor{lightpurple}\textbf{0.906} & \cellcolor{lightpurple}0.907 & \cellcolor{lightpurple}0.905 & \cellcolor{lightpurple}\textbf{0.906} & \cellcolor{lightpurple}\textbf{0.818} & \cellcolor{lightpurple}\textbf{0.910} & \cellcolor{lightpurple}0.707 & \cellcolor{lightpurple}\textbf{0.795} & \cellcolor{lightpurple}\textbf{0.900} & \cellcolor{lightpurple}0.942 & \cellcolor{lightpurple}\textbf{0.852} & \cellcolor{lightpurple}\textbf{0.895} \\ \hline
    
    \multirow{6}{*}{{\textit{Random}}}
    & origin & 0.885 & 0.867 & 0.910 & 0.888 & 0.792 & 0.792 & 0.792 & 0.792 & 0.884 & {0.987} & 0.778 & 0.870 \\
    & VCD & {0.892} & {0.881} & {0.906} & {0.893} & {0.808} & {0.769} & {0.881} & {0.821} & {0.887} & {0.962} & {0.806} & {0.877} \\
    & OPERA & {0.878} & \textbf{0.918} & {0.831} & {0.873} & {0.817} & {0.819} & {0.814} & {0.816} & {0.886} & \textbf{0.990} & {0.780} & {0.872} \\
    & Woodpecker & {0.834} & {0.788} & \textbf{0.914} & {0.846} & 0.818 & 0.766 & \textbf{0.917} & \textbf{0.835} & {0.886} & {0.983} & {0.785} & {0.873} \\
    & VTI & {0.891} & {0.906} & {0.870} & {0.888} & 0.799 & 0.761 & 0.871 & 0.812 & {0.895} & {0.988} & {0.800} & {0.884} \\
    & \cellcolor{lightpurple}Ours & \cellcolor{lightpurple}\textbf{0.897} & \cellcolor{lightpurple}0.890 & \cellcolor{lightpurple}0.905 & \cellcolor{lightpurple}\textbf{0.897} & \cellcolor{lightpurple}\textbf{0.843} & \cellcolor{lightpurple}\textbf{0.972} & \cellcolor{lightpurple}0.705 & \cellcolor{lightpurple}0.818 & \cellcolor{lightpurple}\textbf{0.916} & \cellcolor{lightpurple}0.976 & \cellcolor{lightpurple}\textbf{0.852} & \cellcolor{lightpurple}\textbf{0.910} \\ \hline
    \end{tabular}%
    }
    \caption{Results comparison on the POPE benchmark.}
    
    \label{tab:pope}
    \end{table*}

\section{Experiments}

\begin{figure}[htbp]
\centering
\includegraphics[width=0.7\columnwidth]{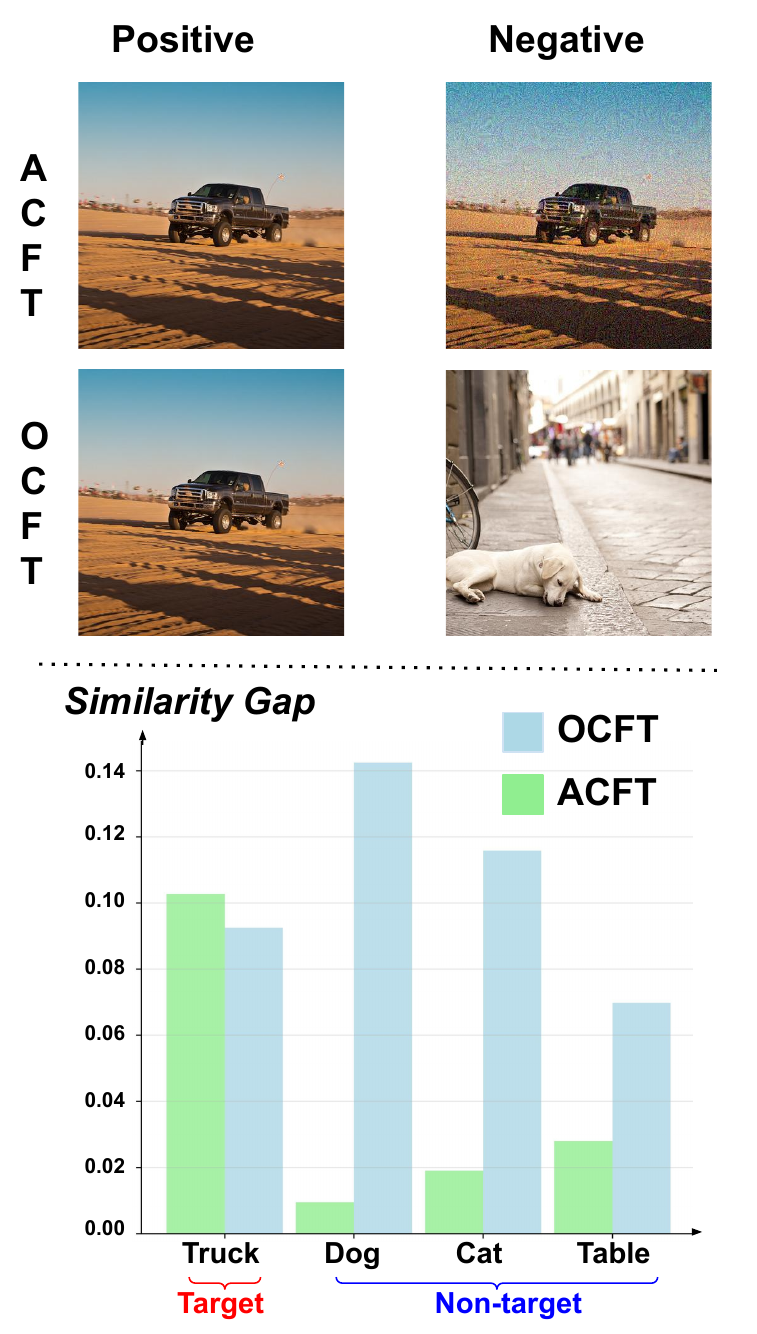}
\Description{Comparison of OCFT and ACFT showing positive-negative sample examples and cosine similarity gaps.}
\caption{Comparison between OCFT and ACFT. \textbf{Top}: Positive and negative samples for ACFT and OCFT (target: ``truck''). \textbf{Bottom}: Cosine similarity gaps for target and non-target objects.}
\label{fig:ocft}
\end{figure}

\subsection{Target MLLMs}

To verify the effectiveness of our method, we adopted three MLLMs as the target models: LLaVA v1.5-7B~\cite{liu2023llava}, MiniGPT-4 13B~\cite{zhu2023minigpt4}, and Qwen2.5-VL-7B~\cite{bai2025qwen25vltechnicalreport}. LLaVA and MiniGPT-4 are widely used as target models in previous research~\cite{leng2023vcd,huang2024opera,yin2024woodpecker,liu2024vti}, and we specifically selected them to ensure a fair comparison with prior work. Qwen2.5-VL is a recent state-of-the-art MLLM that we include to demonstrate the applicability of ACFT to modern architectures. Furthermore, to prove that ACFT generalizes effectively across diverse model families, we extended our evaluation to InternVL-3.5-4B~\cite{wang2025internvl35advancingopensourcemultimodal}. As shown in the SM, ACFT yields consistent performance improvements on this additional architecture.

\subsection{Baseline Methods}

We compared our ACFT method with two categories of baselines. The first category consists of four inference-time methods: two input-level decoding methods, Visual Contrastive Decoding (VCD)~\cite{leng2023vcd} and OPERA~\cite{huang2024opera}; one post-processing method, Woodpecker~\cite{yin2024woodpecker}; and one latent-space-processing method, Visual and Textual Intervention (VTI)~\cite{liu2024reducinghallucinationsvisionlanguagemodels}. The second category consists of four post-training methods: a standard SFT baseline trained with cross-entropy loss on the same data as ACFT, LLaVA-RLHF~\cite{sun2023aligning}, OPA-DPO~\cite{yang2025opadpo}, and CHiP-DPO~\cite{fu2025chip}. Since most post-training baselines only release model weights based on LLaVA v1.5 7B, we conduct the post-training comparison on this backbone for a fair evaluation.

\subsection{Benchmarks}

We evaluated all methods using two benchmarks: POPE \cite{li2023evaluatingobjecthallucinationlarge} and MME \cite{fu2024mmecomprehensiveevaluationbenchmark}.

\textbf{POPE} is a benchmark specifically designed to assess object hallucinations in images. It formulates hallucination assessment as a binary classification task: given an image and an object, the model is asked simple queries "Is there an OBJ in the image?". POPE includes three subsets: \textbf{Random}, \textbf{Popular} and \textbf{Adversarial}. We use the official benchmark of POPE, which includes 3,000 QA pairs for each subset.

\textbf{MME} is a comprehensive benchmark for evaluating MLLMs across 14 tasks spanning perception and cognition. Among them, the \textbf{Existence} subset is the most relevant to object hallucination: it requires models to judge whether a given object or attribute exists in the image, typically answering with "Yes" or "No"— same as POPE.


We selected these two benchmarks because all their questions are answered with either 'Yes' or 'No'. We believe the model's reasoning may rely more on image dimension when the output text is relatively short. This short-text-output setting aligns well with our focus on the image-prior perspective to explore the causes of hallucinations and corresponding mitigation methods in this study. In contrast, many previous research ~\cite{leng2023vcd,zhu2024ibdalleviatinghallucinationslarge} focused on the long-text-output settings (like "describe this image in detail") that align with their language-prior perspective to explore the causes of hallucinations. Although ACFT is targeted on short-answer QAs, it can also reduce object hallucinations in open-ended generation tasks. Evaluations on benchmarks such as CHAIR~\cite{rohrbach2018object} and ObjectHal~\cite{yu2024rlaifv} (detailed in Section~\ref{sec:description_main}) confirm the effectiveness of ACFT.

\begin{table}[htbp]
\centering
\renewcommand{\arraystretch}{1.08}
\resizebox{\columnwidth}{!}{%
\begin{tabular}{cccccc}
\hline
\textbf{Subset} & \textbf{Method} & \textbf{ACC} & \textbf{Precision} & \textbf{Recall} & \textbf{F1 Score} \\ \hline
\multirow{2}{*}{\textit{Adversarial}} & OCFT & 0.483 & 0.489 &  0.784 & 0.602 \\
                                      & \cellcolor{lightpurple} ACFT           & \cellcolor{lightpurple}\textbf{0.841} & \cellcolor{lightpurple}\textbf{0.802} & \cellcolor{lightpurple}\textbf{0.905} & \cellcolor{lightpurple}\textbf{0.850} \\ \hline
\multirow{2}{*}{\textit{Popular}}     & OCFT & 0.832 & 0.867 & 0.784 &  0.824\\
                                      & \cellcolor{lightpurple} ACFT & \cellcolor{lightpurple}\textbf{0.906} & \cellcolor{lightpurple}\textbf{0.907} & \cellcolor{lightpurple}\textbf{0.905} & \cellcolor{lightpurple}\textbf{0.906} \\ \hline
\multirow{2}{*}{\textit{Random}}                       & OCFT & 0.721 & 0.696 &0.784  & 0.737 \\
                                      & \cellcolor{lightpurple}ACFT           & \cellcolor{lightpurple}\textbf{0.897} & \cellcolor{lightpurple}\textbf{0.890} & \cellcolor{lightpurple}\textbf{0.905} & \cellcolor{lightpurple}\textbf{0.897} \\ \hline
\end{tabular}%
}
\caption{Comparison between OCFT and ACFT.}
\label{tab:ocft}
\end{table}

\subsection{Evaluation Metrics}
In our experimental setup, all questions were answered with either 'Yes' or 'No'. Therefore, whether the model hallucinated can be formulated as a classification problem. We chose four widely used classification metrics including \textit{accuracy}, \textit{precision},
\textit{recall}, and \textit{F1 score} for evaluation.

\subsection{Implementation Details}
We present the implementation details of ACFT and baseline methods like adversarial attack settings, finetuning strategies, hyperparameter settings, GPU, etc. in the SM.


\subsection{Why Aligned Pairs Matter: OCFT vs. ACFT} \label{sec:OCFT}


A central claim of this work is that \emph{aligned} contrastive pairs, where positive and negative images differ only in the target object's hallucination-relevant features, are essential for effective hallucination mitigation. We validate this by comparing ACFT against ordinary contrastive fine-tuning (OCFT), which pairs a matched image with a randomly sampled unrelated image.

We use the same 3,000 images from the COCO dataset~\cite{lin2015microsoftcococommonobjects} as the training set. For ACFT, we employ AHAF to generate aligned positive-negative image pairs for contrast finetuning. In contrast, OCFT constructs image pairs by randomly sampling a "positive" image that matches a given text anchor (e.g., ``cat") and a "negative" image drawn arbitrarily from other categories (e.g. ``dog"), resulting in unaligned and semantically inconsistent pairs. Both methods use the same fine-tuning strategy on LLaVA v1.5 7B, and are evaluated on three subsets of the POPE benchmark.

As shown in Table~\ref{tab:ocft}, ACFT significantly outperformed OCFT in mitigating object hallucination, with accuracy improvements of 35.8\%, 7.4\%, and 17.6\% on the three subsets, respectively. Notably, OCFT performed particularly poorly on the \textit{Adversarial} subset, only achieving an accuracy of 0.483, which was even much lower than that of the original LLaVA model. These results highlight the effectiveness of ACFT compared to OCFT.

We computed the similarity gaps between positive and negative samples for the target object and non-target objects, respectively, using Equation \eqref{eq:gap}. Assuming the embeddings of the positive image, negative image, and anchor text are $z^+$, $z^-$, and $t$, respectively, we define the similarity gap $\Delta$ as:
\begin{equation}\label{eq:gap}
\Delta (z^+,z^-,t)
= | \psi(z^+, t)- \psi(z^-, t)|,
\end{equation}
where the cosine similarity $\psi$ between two vectors is defined in Equation \ref{eq:sim}. A representative example is shown in Figure~\ref{fig:ocft}. More experimental details and results are provided in the SM. These results indicate that for ACFT, the similarity gap between positive and negative samples for the target object is clearly distinguishable from that for non-target objects. This property facilitates the model in learning consistent rules to focus on differences in the target object's features, thereby enhancing its ability to distinguish hallucination-inducing samples from non-hallucinating ones. However, OCFT lacks this property, which partially explains why ACFT achieves superior performance compared to OCFT.

\begin{table}[htbp]
\centering
\resizebox{\columnwidth}{!}{%
\begin{tabular}{cccccc}
\hline
\textbf{Model} & \textbf{Method} & \textbf{ACC} & \textbf{Precision} & \textbf{Recall} & \textbf{F1 Score} \\ \hline
\multirow{6}{*}{\textbf{LlaVA v1.5}} & origin     & 0.950 & 0.909 & \textbf{1.000} & 0.952 \\
                                        & VCD        & {0.950} & {0.935} & {0.967} & {0.951} \\
                                        & OPERA      & {0.933} & {0.964} & {0.900} & {0.931} \\
                                        & Woodpecker & {0.933} & {0.882} & {1.000} & {0.937} \\
                                        & VTI        & {0.967}      & {0.967}      &  {0.967}   &      {0.967} \\
                                        & \cellcolor{lightpurple}Ours       & \cellcolor{lightpurple}\textbf{0.983} & \cellcolor{lightpurple}\textbf{1.000} & \cellcolor{lightpurple}0.967 & \cellcolor{lightpurple}\textbf{0.983} \\ \hline
\multirow{6}{*}{\textbf{MiniGPT4}}  & origin     & 0.850 & 0.800 & \textbf{0.933} & 0.861\\
                                        & VCD     &  {0.867} & {0.867} & {0.867} & {0.867} \\
                                        & OPERA      &  {0.850}    &  {0.838}    &  {0.867}    &  {0.852}   \\
                                        & Woodpecker &   {0.833}    &   {0.794}    &   {0.900}    &   {0.843}    \\
                                        & VTI        &   {0.883}   &  {0.896}    &  {0.867}     &    {0.881}  \\
                                        & \cellcolor{lightpurple}Ours       & \cellcolor{lightpurple}\textbf{0.900} & \cellcolor{lightpurple}\textbf{0.900} & \cellcolor{lightpurple}0.900 & \cellcolor{lightpurple}\textbf{0.900} \\ \hline
\multirow{6}{*}{\textbf{Qwen2.5-VL}}  & origin     & \textbf{1.000} & \textbf{1.000} & \textbf{1.000} & \textbf{1.000} \\
                                        & VCD        & \textbf{1.000}& \textbf{1.000} & \textbf{1.000} & \textbf{1.000} \\
                                        & OPERA      & \textbf{1.000} & \textbf{1.000} & \textbf{1.000} & \textbf{1.000}\\
                                        & Woodpecker & 0.967 & 0.967 & 0.967 & 0.967 \\
                                        & VTI        & \textbf{1.000} & \textbf{1.000} & \textbf{1.000} & \textbf{1.000}\\
                                        & \cellcolor{lightpurple}Ours       & \cellcolor{lightpurple}\textbf{1.000} & \cellcolor{lightpurple}\textbf{1.000} & \cellcolor{lightpurple}\textbf{1.000} & \cellcolor{lightpurple}\textbf{1.000} \\ \hline
\end{tabular}%
}
\caption{Results comparison on the MME Existence subset.}
\label{tab:mmeexist}
\end{table}

\subsection{Effectiveness of ACFT on Mitigating Hallucination}
We compared ACFT with other baseline methods on POPE and MME-Existence benchmarks across three target MLLMs to show its effectiveness on mitigating hallucination.

\begin{figure*}[htbp]
\centering
\includegraphics[width=0.7\textwidth]{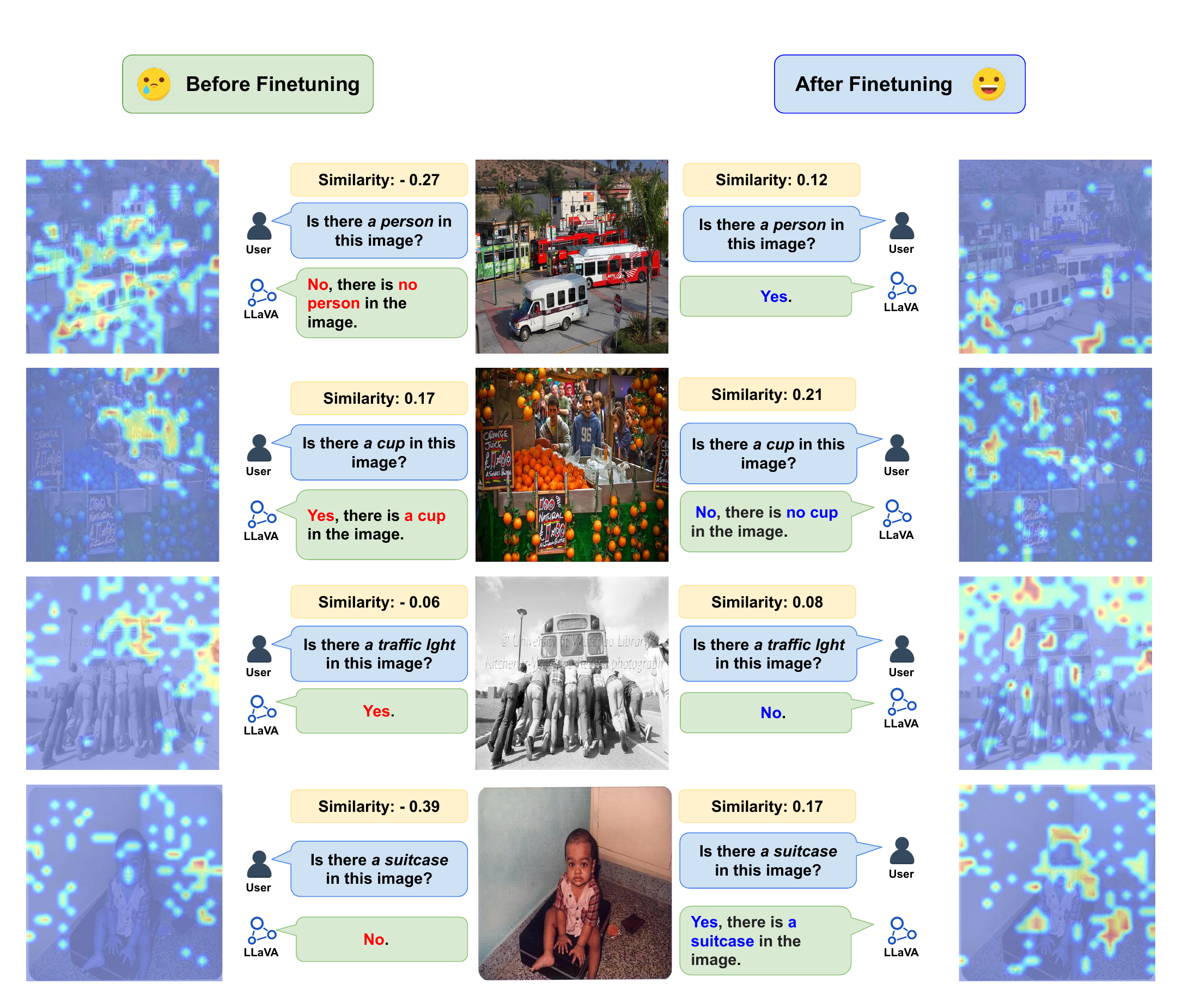}
\vspace{-6pt}
\Description{Detailed visualization showing Grad-CAM attention maps and cosine similarity improvements after ACFT.}
\caption{Visualization results of ACFT. ACFT improved cosine similarity between image and text embeddings, while Grad-CAM attention maps became more focused on target objects. \textcolor{blue}{Blue} = correct responses, \textcolor{red}{red} = hallucinations.}
\label{fig:visual}
\end{figure*}

\subsubsection{Evaluation on POPE} Table \ref{tab:pope} shows that ACFT significantly outperforms all baseline methods across three different target MLLMs. For the LlaVA model, ACFT achieved accuracies of 0.841, 0.906, and 0.897 on the three subsets of POPE, surpassing the second-best baseline by 3.3\%, 2.0\%, and 0.5\%, respectively. Similarly, for the MiniGPT-4 model, ACFT achieved the best accuracy, surpassing the second-best baseline by 3.0\%, 5.3\%, and 2.5\%, respectively. For the Qwen2.5-VL model, despite its already strong baseline performance, ACFT still consistently improves accuracy across all three subsets (from 0.864 to 0.877, 0.875 to 0.900, and 0.884 to 0.916), demonstrating ACFT's effectiveness even on modern, high-performing MLLMs.

The advantage stems from ACFT's contrastive pairs during training. These pairs enhance the alignment between visual and textual embedding and encourage the visual modality to focus on the correct image features. In contrast, existing methods are often constrained by their design: VCD and OPERA rely heavily on language priors or text-side decoding, limiting their effectiveness against image-induced hallucinations. Meanwhile, VTI's inference-only adjustments lack generalization, and Woodpecker is susceptible to errors from external grounding modules. These limitations hinder baseline performance on vision-dependent tasks, whereas ACFT's refined visual alignment yields superior results.

\subsubsection{Evaluation on MME-Existence}

The results in Table~\ref{tab:mmeexist} showed that ACFT achieved the best performance on the MME-Existence subset. For the LLaVA model, although the original version already attained a relatively high accuracy of 0.950, ACFT further improved it by 3.3\%. In contrast, other baseline methods such as OPERA and Woodpecker, even degraded the model's performance. For the MiniGPT4 model, ACFT brought an improvement, 1.7\%, compared to the best-performing baseline. For the Qwen2.5-VL model, both the original model and ACFT achieve perfect accuracy of 1.000, reflecting that the MME-Existence task is relatively easy for modern VLMs, leaving no headroom for further improvement. These results highlight the robustness and effectiveness of ACFT across different target models and benchmarks.

\begin{table}[t]
\centering
\resizebox{\columnwidth}{!}{%
\begin{tabular}{cccccc}
\hline
\textbf{Model} & \textbf{Method} & \textbf{ACC} & \textbf{Precision} & \textbf{Recall} & \textbf{F1 Score} \\ \hline
\multirow{2}{*}{\textbf{LlaVA v1.5}} & Origin & 0.728 & 0.666 & 0.916 & 0.771 \\
                                        &\cellcolor{lightpurple} Ours   & \cellcolor{lightpurple}\textbf{0.747} & \cellcolor{lightpurple}\textbf{0.683} & \cellcolor{lightpurple}\textbf{0.922} & \cellcolor{lightpurple}\textbf{0.785} \\ \hline
\multirow{2}{*}{\textbf{MiniGPT4}}  & Origin & 0.538 & 0.531 & 0.655 & 0.586 \\
                                        & \cellcolor{lightpurple}Ours   & \cellcolor{lightpurple}\textbf{0.548} & \cellcolor{lightpurple}\textbf{0.538} & \cellcolor{lightpurple}\textbf{0.672} & \cellcolor{lightpurple}\textbf{0.598} \\ \hline
\multirow{2}{*}{\textbf{Qwen2.5-VL}} & Origin & 0.870 & 0.836 & \textbf{0.920} & \textbf{0.878} \\
                                        & \cellcolor{lightpurple}Ours   & \cellcolor{lightpurple}\textbf{0.874} & \cellcolor{lightpurple}\textbf{0.928} & \cellcolor{lightpurple}0.810 & \cellcolor{lightpurple}0.865 \\ \hline
\end{tabular}%
}
\caption{Results comparison on the MME whole benchmark.}
\label{tab:mmewhole}
\end{table}

\subsubsection{Comparison with Post-training Methods}

We further compare ACFT with post-training baselines on LLaVA v1.5 7B, since most post-training methods only release weights for this backbone. As shown in Table~\ref{tab:pope_mme_results}, under a comparable data budget ($\sim$6k samples), ACFT achieves the best overall performance. This indicates that the gains of ACFT stem from the adversarially constructed contrastive pairs and the contrastive loss, which more directly target the visual misalignment underlying object hallucinations, rather than simply from additional fine-tuning.

\begin{table}[htbp]
\centering
\resizebox{\columnwidth}{!}{%
\begin{tabular}{ccccccc}
\toprule
\textbf{Bench.} & \textbf{Subset} & \textbf{Method} & \textbf{ACC} & \textbf{Pre.} & \textbf{Rec.} & \textbf{F1} \\
\midrule
\multirow{15}{*}{POPE}
  & \multirow{5}{*}{\textit{Adv.}}
      & SFT                & 0.797 & 0.743 & \textbf{0.906} & 0.817 \\
  &   & LLaVA-RLHF         & 0.813 & 0.835 & 0.780          & 0.806 \\
  &   & CHiP-DPO           & 0.839 & 0.923 & 0.739          & 0.821 \\
  &   & OPA-DPO             & 0.827 & \textbf{0.944} & 0.697 & 0.801 \\
  &   & \cellcolor{lightpurple}ACFT & \cellcolor{lightpurple}\textbf{0.841} & \cellcolor{lightpurple}0.802 & \cellcolor{lightpurple}0.905 & \cellcolor{lightpurple}\textbf{0.850} \\
\cmidrule(lr){2-7}
  & \multirow{5}{*}{\textit{Pop.}}
      & SFT                & 0.862 & 0.832 & \textbf{0.906} & 0.867 \\
  &   & LLaVA-RLHF         & 0.847 & 0.901 & 0.780          & 0.836 \\
  &   & CHiP-DPO           & 0.855 & 0.962 & 0.738          & 0.835 \\
  &   & OPA-DPO             & 0.840 & \textbf{0.975} & 0.698 & 0.813 \\
  &   & \cellcolor{lightpurple}ACFT & \cellcolor{lightpurple}\textbf{0.906} & \cellcolor{lightpurple}0.907 & \cellcolor{lightpurple}0.905 & \cellcolor{lightpurple}\textbf{0.906} \\
\cmidrule(lr){2-7}
  & \multirow{5}{*}{\textit{Rand.}}
      & SFT                & 0.896 & 0.888 & \textbf{0.906} & \textbf{0.897} \\
  &   & LLaVA-RLHF         & 0.867 & 0.943 & 0.780          & 0.854 \\
  &   & CHiP-DPO           & 0.863 & 0.984 & 0.739          & 0.844 \\
  &   & OPA-DPO             & 0.845 & \textbf{0.991} & 0.696 & 0.818 \\
  &   & \cellcolor{lightpurple}ACFT & \cellcolor{lightpurple}\textbf{0.897} & \cellcolor{lightpurple}0.890 & \cellcolor{lightpurple}0.905 & \cellcolor{lightpurple}\textbf{0.897} \\
\midrule
\multirow{10}{*}{MME}
  & \multirow{5}{*}{\textit{Exist.}}
      & SFT                & 0.950 & 0.935 & \textbf{0.967} & 0.951 \\
  &   & LLaVA-RLHF         & 0.967 & 0.967 & \textbf{0.967} & 0.967 \\
  &   & CHiP-DPO           & 0.967 & \textbf{1.000} & 0.933 & 0.965 \\
  &   & OPA-DPO             & 0.967 & \textbf{1.000} & 0.933 & 0.965 \\
  &   & \cellcolor{lightpurple}ACFT & \cellcolor{lightpurple}\textbf{0.983} & \cellcolor{lightpurple}\textbf{1.000} & \cellcolor{lightpurple}\textbf{0.967} & \cellcolor{lightpurple}\textbf{0.983} \\
\cmidrule(lr){2-7}
  & \multirow{5}{*}{\textit{Whole}}
      & SFT                & 0.736 & 0.718 & 0.778          & 0.747 \\
  &   & LLaVA-RLHF         & 0.717 & 0.800 & 0.578          & 0.671 \\
  &   & CHiP-DPO           & 0.653 & \textbf{0.925} & 0.334 & 0.491 \\
  &   & OPA-DPO             & \textbf{0.753} & 0.897 & 0.572 & 0.699 \\
  &   & \cellcolor{lightpurple}ACFT & \cellcolor{lightpurple}0.747 & \cellcolor{lightpurple}0.683 & \cellcolor{lightpurple}\textbf{0.922} & \cellcolor{lightpurple}\textbf{0.785} \\
\bottomrule
\end{tabular}%
}
\caption{Post-training baselines vs. ACFT on LLaVA v1.5 7B ($\sim$6k samples).}
\label{tab:pope_mme_results}
\end{table}

\vspace{-2pt}
\subsection{Generalization to Description-level Tasks}
\label{sec:description_main}

Although ACFT is trained on short-answer data, we evaluate whether it generalizes to open-ended description tasks. We conduct experiments on four description-level benchmarks: CHAIR~\cite{rohrbach2018object}, CCEval~\cite{zhai2023halleswitch}, AMBERA~\cite{wang2023llm}, and ObjectHal~\cite{yu2024rlaifv}, all on LLaVA v1.5-7B. As shown in Table~\ref{tab:hallucination_caption_benchmarks}, ACFT consistently reduces caption-level hallucination across all four benchmarks, with lower CHAIR scores and hallucination rates compared to the original model. This supports our claim that strengthening visual perception and multimodal alignment benefits not only binary settings but also transfers to open-ended description tasks. Although the absolute gains are smaller than those on binary benchmarks---unsurprising given that long-form generation is more affected by language priors---these results demonstrate that ACFT provides robust visual grounding that generalizes beyond its training distribution.

\begin{table}[t]
\centering
\renewcommand{\arraystretch}{1.08}
\small
\begin{tabular}{cccc}
\hline
\textbf{Benchmark} & \textbf{Metric} & \textbf{Original} & \cellcolor{lightpurple}\textbf{ACFT} \\ \hline
\multirow{2}{*}{CHAIR}
    & CHAIR$_s$ ${\downarrow}$          & 0.508  & \cellcolor{lightpurple}\textbf{0.494} \\
    & CHAIR$_i$ ${\downarrow}$       & 0.142  & \cellcolor{lightpurple}\textbf{0.137} \\ \hline
\multirow{2}{*}{CCEval}
    & CHAIR$_s$ ${\downarrow}$           & 0.870  & \cellcolor{lightpurple}\textbf{0.850} \\
    & CHAIR$_i$ ${\downarrow}$           & 0.348  & \cellcolor{lightpurple}\textbf{0.328} \\ \hline
\multirow{4}{*}{\shortstack{AMBERA\\Generative}}
    & CHAIR ${\downarrow}$                & 0.112   & \cellcolor{lightpurple}\textbf{0.089}   \\
    & Hal ${\downarrow}$                  & 0.488   & \cellcolor{lightpurple}\textbf{0.483}  \\
    & Cover ${\uparrow}$               & \textbf{0.518}   & \cellcolor{lightpurple}0.513  \\
    & Cog ${\downarrow}$                  & 0.047    & \cellcolor{lightpurple}\textbf{0.043}   \\ \hline
\multirow{4}{*}{\shortstack{AMBERA\\Discrim.}}
    & ACC ${\uparrow}$                  & 0.716   & \cellcolor{lightpurple}\textbf{0.729}  \\
    & Precision ${\uparrow}$            & 0.933   & \cellcolor{lightpurple}\textbf{0.941}  \\
    & Recall ${\uparrow}$               & 0.617   & \cellcolor{lightpurple}\textbf{0.630}  \\
    & F1 ${\uparrow}$                   & 0.743   & \cellcolor{lightpurple}\textbf{0.755}  \\ \hline
\multirow{2}{*}{ObjectHal}
    & Resp. Hal ${\downarrow}$         & 0.568  & \cellcolor{lightpurple}\textbf{0.562} \\
    & Obj Hal ${\downarrow}$              & 0.283  & \cellcolor{lightpurple}\textbf{0.277} \\ \hline
\end{tabular}
\caption{Comparison on description-level hallucination benchmarks (LLaVA v1.5 7B).}
\label{tab:hallucination_caption_benchmarks}
\end{table}

\vspace{-6pt}
\subsection{Effectiveness of ACFT on Visual Comprehension}
Prior research \cite{liu2024automaticallygeneratingvisualhallucination} suggests that fine-tuning on specific tasks may compromise a model's general capabilities. To address whether ACFT impairs general visual comprehension while mitigating hallucinations, we evaluated it on the full MME benchmark. As shown in Table~\ref{tab:mmewhole}, ACFT does not degrade performance; rather, it yields marginal improvements in overall scores for LLaVA and MiniGPT4, and a comparable score for Qwen2.5-VL. These results indicate that ACFT not only mitigates hallucinations but also preserves the model's robustness in visual feature extraction and comprehension, thereby maintaining or improving its general visual understanding ability.


\vspace{-3pt}

\subsection{Visualization}
Figure~\ref{fig:visual} visualizes how ACFT corrects the visual-origin hallucination signatures identified in Section~\ref{intro}. After ACFT, cosine similarity between image and text embeddings improves markedly, and Grad-CAM attention shifts toward the semantically appropriate distribution: in Cases 1, 4 (object present), attention concentrates on the correct region; in Cases 2, 3 (object absent), it becomes properly dispersed. Entropy analysis confirms the correction: for present objects, entropy decreases by 8.7\% and 2.6\%; for absent objects, it increases by 7.4\% and 6.4\%. These results suggest that ACFT effectively addresses the two diagnostic signatures, embedding misalignment and inverted attention, by improving the underlying visual representations.

\section{Conclusion}

This paper challenges the prevailing language-prior explanation of object hallucination in MLLMs by identifying and systematically characterizing \emph{visual-origin hallucination}, a distinct mechanism driven by incorrect visual feature extraction and image-text embedding misalignment. We provide quantitative evidence through cosine similarity analysis and attention entropy measurements, showing that hallucinated samples exhibit systematically lower cross-modal alignment and inverted attention patterns. Guided by this diagnosis, we propose AHAF, which serves as both a diagnostic probe revealing the fragility of MLLM visual representations and an efficient generator of aligned contrastive training pairs, and ACFT, a data-efficient contrastive fine-tuning method requiring only 0.9\% of COCO data with zero inference overhead. Extensive experiments on POPE, MME, and four description-level benchmarks demonstrate state-of-the-art performance, and visualization analysis suggests that ACFT effectively mitigates the two identified hallucination signatures by improving the underlying visual representations.

\clearpage
\begin{acks}
This work was supported by the National Natural Science Foundation of China (No.U2341228 and No.62576187).
\end{acks}

\bibliographystyle{ACM-Reference-Format}
\bibliography{main,sm_references}

@misc{liu2024reducinghallucinationsvisionlanguagemodels,
      title={Reducing Hallucinations in Vision-Language Models via Latent Space Steering}, 
      author={Sheng Liu and Haotian Ye and Lei Xing and James Zou},
      year={2024},
      eprint={2410.15778},
      archivePrefix={arXiv},
      primaryClass={cs.CV},
      url={https://arxiv.org/abs/2410.15778}, 
}

@misc{madry2019deeplearningmodelsresistant,
      title={Towards Deep Learning Models Resistant to Adversarial Attacks}, 
      author={Aleksander Madry and Aleksandar Makelov and Ludwig Schmidt and Dimitris Tsipras and Adrian Vladu},
      year={2019},
      eprint={1706.06083},
      archivePrefix={arXiv},
      primaryClass={stat.ML},
      url={https://arxiv.org/abs/1706.06083}, 
}

@article{liu2023llava, title={{Visual Instruction Tuning}}, author={Liu, Haotian and Li, Chunyuan and Wu, Qingyang and Lee, Yong Jae}, journal={arXiv preprint arXiv:2304.08485}, year={2023} }

@article{zhu2023minigpt4, title={MiniGPT-4: Enhancing Vision-Language Understanding with Advanced Large Language Models}, author={Zhu, Deyao and Chen, Jun and Shen, Xiaoqian and Li, Xiang and Elhoseiny, Mohamed}, journal={arXiv preprint arXiv:2304.10592}, year={2023} }

@article{leng2023vcd,
  title={Mitigating Object Hallucinations in Large Vision-Language Models through Visual Contrastive Decoding},
  author={Leng, Sicong and Zhang, Hang and Chen, Guanzheng and Li, Xin and Lu, Shijian and Miao, Chunyan and Bing, Lidong},
  journal={arXiv preprint arXiv:2311.16922},
  year={2023}
}

@article{huang2024opera,
  title={OPERA: Alleviating Hallucination in Multi-Modal Large Language Models via Over-Trust Penalty and Retrospection-Allocation},
  author={Huang, Qidong and Dong, Xiaoyi and Zhang, Pan and Wang, Bin and He, Conghui and Wang, Jiaqi and Lin, Dahua and Zhang, Weiming and Yu, Nenghai},
  journal={arXiv preprint arXiv:2311.17911},
  year={2024}
}

@article{yin2024woodpecker,
  title={Woodpecker: Hallucination Correction for Multimodal Large Language Models},
  author={Yin, Shukang and Fu, Chaoyou and Zhao, Sirui and Xu, Tong and Wang, Hao and Sui, Dianbo and Shen, Yunhang and Li, Ke and Sun, Xing and Chen, Enhong},
  journal={arXiv preprint arXiv:2310.16045},
  year={2024}
}

@misc{li2023evaluatingobjecthallucinationlarge,
      title={Evaluating Object Hallucination in Large Vision-Language Models}, 
      author={Yifan Li and Yifan Du and Kun Zhou and Jinpeng Wang and Wayne Xin Zhao and Ji-Rong Wen},
      year={2023},
      eprint={2305.10355},
      archivePrefix={arXiv},
      primaryClass={cs.CV},
      url={https://arxiv.org/abs/2305.10355}, 
}

@misc{fu2024mmecomprehensiveevaluationbenchmark,
      title={MME: A Comprehensive Evaluation Benchmark for Multimodal Large Language Models}, 
      author={Chaoyou Fu and Peixian Chen and Yunhang Shen and Yulei Qin and Mengdan Zhang and Xu Lin and Jinrui Yang and Xiawu Zheng and Ke Li and Xing Sun and Yunsheng Wu and Rongrong Ji},
      year={2024},
      eprint={2306.13394},
      archivePrefix={arXiv},
      primaryClass={cs.CV},
      url={https://arxiv.org/abs/2306.13394}, 
}

@misc{lin2015microsoftcococommonobjects,
      title={Microsoft COCO: Common Objects in Context}, 
      author={Tsung-Yi Lin and Michael Maire and Serge Belongie and Lubomir Bourdev and Ross Girshick and James Hays and Pietro Perona and Deva Ramanan and C. Lawrence Zitnick and Piotr Dollár},
      year={2015},
      eprint={1405.0312},
      archivePrefix={arXiv},
      primaryClass={cs.CV},
      url={https://arxiv.org/abs/1405.0312}, 
}

@misc{zhu2024ibdalleviatinghallucinationslarge,
      title={IBD: Alleviating Hallucinations in Large Vision-Language Models via Image-Biased Decoding}, 
      author={Lanyun Zhu and Deyi Ji and Tianrun Chen and Peng Xu and Jieping Ye and Jun Liu},
      year={2024},
      eprint={2402.18476},
      archivePrefix={arXiv},
      primaryClass={cs.CV},
      url={https://arxiv.org/abs/2402.18476}, 
}

@misc{liu2024automaticallygeneratingvisualhallucination,
      title={Automatically Generating Visual Hallucination Test Cases for Multimodal Large Language Models}, 
      author={Zhongye Liu and Hongbin Liu and Yuepeng Hu and Zedian Shao and Neil Zhenqiang Gong},
      year={2024},
      eprint={2410.11242},
      archivePrefix={arXiv},
      primaryClass={cs.CV},
      url={https://arxiv.org/abs/2410.11242}, 
}

@misc{zhang2024redundancyrelevanceinformationflow,
      title={From Redundancy to Relevance: Information Flow in LVLMs Across Reasoning Tasks}, 
      author={Xiaofeng Zhang and Yihao Quan and Chen Shen and Xiaosong Yuan and Shaotian Yan and Liang Xie and Wenxiao Wang and Chaochen Gu and Hao Tang and Jieping Ye},
      year={2024},
      eprint={2406.06579},
      archivePrefix={arXiv},
      primaryClass={cs.CL},
      url={https://arxiv.org/abs/2406.06579}, 
}

@article{openai2023gpt4, title={{GPT-4} Technical Report}, author={OpenAI}, journal={arXiv preprint arXiv:2303.08774}, year={2023} }

@inproceedings{rohrbach2018object, title={Object Hallucination in Image Captioning}, author={Rohrbach, Anna and Hendricks, Lisa Anne and Burns, Kaylee and Darrell, Trevor and Saenko, Kate}, booktitle={European Conference on Computer Vision (ECCV)}, pages={635--651}, year={2018} }

@inproceedings{radford2021clip, title={Learning Transferable Visual Models from Natural Language Supervision}, author={Radford, Alec and Kim, Jong Wook and Hallacy, Chris and Ramesh, Aditya and Goh, Gabriel and Agarwal, Sandhini and \emph{et al.}}, booktitle={International Conference on Machine Learning (ICML) Workshop}, year={2021} }

@inproceedings{wei2024lehace, title={Toward a Stable, Fair, and Comprehensive Evaluation of Object Hallucination in Large Vision-Language Models}, author={Wei, Hongliang and Wang, Xingtao and Zhang, Xianqi and Fan, Xiaopeng and Zhao, Debin}, booktitle={Advances in Neural Information Processing Systems (NeurIPS) -- Poster}, year={2024} }

@inproceedings{wu2024logic,
  title={Logical Closed Loop: Uncovering Object Hallucinations in Large Vision-Language Models},
  author={Wu, Junfei and Liu, Qiang and Wang, Ding and Zhang, Jinghao and Wu, Shu and Wang, Liang and Tan, Tieniu},
  booktitle={Findings of the Association for Computational Linguistics: ACL 2024},
  year={2024}
}

@article{sun2025cause,
  title={Exploring Causes and Mitigation of Hallucinations in Large Vision Language Models},
  author={Sun, Yaqi and Atarashi, Kyohei and Takeuchi, Koh and Kashima, Hisashi},
  journal={arXiv preprint arXiv:2502.16842},
  year={2025}
}

@article{liu2024vti,
  title={Reducing Hallucinations in Vision-Language Models via Latent Space Steering},
  author={Liu, Sheng and Ye, Haotian and Xing, Lei and Zou, James},
  journal={arXiv preprint arXiv:2410.15778},
  year={2024}
}

@article{zhu2024ibd,
  title={IBD: Alleviating Hallucinations in Large Vision-Language Models via Image-Biased Decoding},
  author={Zhu, Lanyun and Ji, Deyi and Chen, Tianrun and Xu, Peng and Ye, Jieping and Liu, Jun},
  journal={arXiv preprint arXiv:2402.18476},
  year={2024}
}

@article{jiang2024hacl,
  title={Hallucination Augmented Contrastive Learning for Multimodal Large Language Model},
  author={Jiang, Chaoya and Xu, Haiyang and Dong, Mengfan and Chen, Jiaxing and Ye, Wei and Yan, Ming and Ye, Qinghao and Zhang, Ji and Huang, Fei and Zhang, Shikun},
  journal={arXiv preprint arXiv:2312.06968},
  year={2024}
}

@misc{omeiza2019smoothgradcamenhancedinference,
      title={Smooth Grad-CAM++: An Enhanced Inference Level Visualization Technique for Deep Convolutional Neural Network Models}, 
      author={Daniel Omeiza and Skyler Speakman and Celia Cintas and Komminist Weldermariam},
      year={2019},
      eprint={1908.01224},
      archivePrefix={arXiv},
      primaryClass={cs.CV},
      url={https://arxiv.org/abs/1908.01224}, 
}

@misc{chen2025perturbollavareducingmultimodalhallucinations,
      title={PerturboLLaVA: Reducing Multimodal Hallucinations with Perturbative Visual Training}, 
      author={Cong Chen and Mingyu Liu and Chenchen Jing and Yizhou Zhou and Fengyun Rao and Hao Chen and Bo Zhang and Chunhua Shen},
      year={2025},
      eprint={2503.06486},
      archivePrefix={arXiv},
      primaryClass={cs.CV},
      url={https://arxiv.org/abs/2503.06486}, 
}

@misc{zhou2024analyzingmitigatingobjecthallucination,
      title={Analyzing and Mitigating Object Hallucination in Large Vision-Language Models}, 
      author={Yiyang Zhou and Chenhang Cui and Jaehong Yoon and Linjun Zhang and Zhun Deng and Chelsea Finn and Mohit Bansal and Huaxiu Yao},
      year={2024},
      eprint={2310.00754},
      archivePrefix={arXiv},
      primaryClass={cs.LG},
      url={https://arxiv.org/abs/2310.00754}, 
}

@inproceedings{
yang2025mitigating,
title={Mitigating Hallucination in Large Vision-Language Models via Modular Attribution and Intervention},
author={Tianyun Yang and Ziniu Li and Juan Cao and Chang Xu},
booktitle={The Thirteenth International Conference on Learning Representations},
year={2025},
url={https://openreview.net/forum?id=Bjq4W7P2Us}
}

@misc{zhang2024reflectiveinstructiontuningmitigating,
      title={Reflective Instruction Tuning: Mitigating Hallucinations in Large Vision-Language Models}, 
      author={Jinrui Zhang and Teng Wang and Haigang Zhang and Ping Lu and Feng Zheng},
      year={2024},
      eprint={2407.11422},
      archivePrefix={arXiv},
      primaryClass={cs.CV},
      url={https://arxiv.org/abs/2407.11422}, 
}

@inproceedings{
madry2018towards,
title={Towards Deep Learning Models Resistant to Adversarial Attacks},
author={Aleksander Madry and Aleksandar Makelov and Ludwig Schmidt and Dimitris Tsipras and Adrian Vladu},
booktitle={International Conference on Learning Representations},
year={2018},
url={https://openreview.net/forum?id=rJzIBfZAb},
}

@misc{bai2021recentadvancesadversarialtraining,
      title={Recent Advances in Adversarial Training for Adversarial Robustness}, 
      author={Tao Bai and Jinqi Luo and Jun Zhao and Bihan Wen and Qian Wang},
      year={2021},
      eprint={2102.01356},
      archivePrefix={arXiv},
      primaryClass={cs.LG},
      url={https://arxiv.org/abs/2102.01356}, 
}

@inproceedings{li2024partimagenet++,
  title={Partimagenet++ dataset: Scaling up part-based models for robust recognition},
  author={Li, Xiao and Liu, Yining and Dong, Na and Qin, Sitian and Hu, Xiaolin},
  booktitle={European Conference on Computer Vision},
  pages={396--414},
  year={2024},
  organization={Springer}
}

@article{li2023recognizing,
  title={Recognizing object by components with human prior knowledge enhances adversarial robustness of deep neural networks},
  author={Li, Xiao and Wang, Ziqi and Zhang, Bo and Sun, Fuchun and Hu, Xiaolin},
  journal={IEEE Transactions on Pattern Analysis and Machine Intelligence},
  volume={45},
  number={7},
  pages={8861--8873},
  year={2023},
  publisher={IEEE}
}

@article{liu2025comprehensive,
  title={A comprehensive study on robustness of image classification models: Benchmarking and rethinking},
  author={Liu, Chang and Dong, Yinpeng and Xiang, Wenzhao and Yang, Xiao and Su, Hang and Zhu, Jun and Chen, Yuefeng and He, Yuan and Xue, Hui and Zheng, Shibao},
  journal={International Journal of Computer Vision},
  volume={133},
  number={2},
  pages={567--589},
  year={2025},
  publisher={Springer}
}

@misc{zhai2023halleswitch,
      title={HallE-Switch: Controlling Object Hallucination in Large Vision Language Models}, 
      author={Bohan Zhai and Shijia Yang and Chenfeng Xu and Sheng Shen and Kurt Keutzer and Manling Li},
      year={2023},
      eprint={2310.01779},
      archivePrefix={arXiv},
      primaryClass={cs.CV}
}

@article{wang2023llm,
  title={An LLM-free Multi-dimensional Benchmark for MLLMs Hallucination Evaluation},
  author={Wang, Junyang and Wang, Yuhang and Xu, Guohai and Zhang, Jing and Gu, Yukai and Jia, Haitao and Yan, Ming and Zhang, Ji and Sang, Jitao},
  journal={arXiv preprint arXiv:2311.07397},
  year={2023}
}

@article{sun2023aligning,
  title={Aligning large multimodal models with factually augmented rlhf},
  author={Sun, Zhiqing and Shen, Sheng and Cao, Shengcao and Liu, Haotian and Li, Chunyuan and Shen, Yikang and Gan, Chuang and Gui, Liang-Yan and Wang, Yu-Xiong and Yang, Yiming and others},
  journal={arXiv preprint arXiv:2309.14525},
  year={2023}
}

@article{yu2024rlaifv,
  title={RLAIF-V: Aligning MLLMs through Open-Source AI Feedback for Super GPT-4V Trustworthiness}, 
  author={Yu, Tianyu and Zhang, Haoye and Yao, Yuan and Dang, Yunkai and Chen, Da and Lu, Xiaoman and Cui, Ganqu and He, Taiwen and Liu, Zhiyuan and Chua, Tat-Seng and Sun, Maosong},
  journal={arXiv preprint arXiv:2405.17220},
  year={2024},
}

@inproceedings{fu2025chip,
  title={CHiP: Cross-modal Hierarchical Direct Preference Optimization for Multimodal LLMs},
  author={Fu, Jinlan and Huangfu, Shenzhen and Fei, Hao and Shen, Xiaoyu and Hooi, Bryan and Qiu, Xipeng and Ng, See-Kiong},
  journal={Proceedings of the International Conference on Learning Representations},
  year={2025}
}

@article{yang2025opadpo,
  title={Mitigating Hallucinations in Large Vision-Language Models via DPO: On-Policy Data Hold the Key},
  author={Yang, Zhihe and Luo, Xufang and Han, Dongqi and Xu, Yunjian and Li, Dongsheng},
  journal={arXiv preprint arXiv:2501.09695},
  year={2025}
}

@misc{bai2025qwen25vltechnicalreport,
      title={Qwen2.5-VL Technical Report}, 
      author={Shuai Bai and Keqin Chen and Xuejing Liu and Jialin Wang and Wenbin Ge and Sibo Song and Kai Dang and Peng Wang and Shijie Wang and Jun Tang and Humen Zhong and Yuanzhi Zhu and Mingkun Yang and Zhaohai Li and Jianqiang Wan and Pengfei Wang and Wei Ding and Zheren Fu and Yiheng Xu and Jiabo Ye and Xi Zhang and Tianbao Xie and Zesen Cheng and Hang Zhang and Zhibo Yang and Haiyang Xu and Junyang Lin},
      year={2025},
      eprint={2502.13923},
      archivePrefix={arXiv},
      primaryClass={cs.CV},
      url={https://arxiv.org/abs/2502.13923}, 
}

@misc{wang2025internvl35advancingopensourcemultimodal,
      title={InternVL3.5: Advancing Open-Source Multimodal Models in Versatility, Reasoning, and Efficiency}, 
      author={Weiyun Wang and Zhangwei Gao and Lixin Gu and Hengjun Pu and Long Cui and Xingguang Wei and Zhaoyang Liu and Linglin Jing and Shenglong Ye and Jie Shao and Zhaokai Wang and Zhe Chen and Hongjie Zhang and Ganlin Yang and Haomin Wang and Qi Wei and Jinhui Yin and Wenhao Li and Erfei Cui and Guanzhou Chen and Zichen Ding and Changyao Tian and Zhenyu Wu and Jingjing Xie and Zehao Li and Bowen Yang and Yuchen Duan and Xuehui Wang and Zhi Hou and Haoran Hao and Tianyi Zhang and Songze Li and Xiangyu Zhao and Haodong Duan and Nianchen Deng and Bin Fu and Yinan He and Yi Wang and Conghui He and Botian Shi and Junjun He and Yingtong Xiong and Han Lv and Lijun Wu and Wenqi Shao and Kaipeng Zhang and Huipeng Deng and Biqing Qi and Jiaye Ge and Qipeng Guo and Wenwei Zhang and Songyang Zhang and Maosong Cao and Junyao Lin and Kexian Tang and Jianfei Gao and Haian Huang and Yuzhe Gu and Chengqi Lyu and Huanze Tang and Rui Wang and Haijun Lv and Wanli Ouyang and Limin Wang and Min Dou and Xizhou Zhu and Tong Lu and Dahua Lin and Jifeng Dai and Weijie Su and Bowen Zhou and Kai Chen and Yu Qiao and Wenhai Wang and Gen Luo},
      year={2025},
      eprint={2508.18265},
      archivePrefix={arXiv},
      primaryClass={cs.CV},
      url={https://arxiv.org/abs/2508.18265}, 
}

@misc{liu2024groundingdinomarryingdino,
      title={Grounding DINO: Marrying DINO with Grounded Pre-Training for Open-Set Object Detection},
      author={Shilong Liu and Zhaoyang Zeng and Tianhe Ren and Feng Li and Hao Zhang and Jie Yang and Qing Jiang and Chunyuan Li and Jianwei Yang and Hang Su and Jun Zhu and Lei Zhang},
      year={2024},
      eprint={2303.05499},
      archivePrefix={arXiv},
      primaryClass={cs.CV},
      url={https://arxiv.org/abs/2303.05499},
}

@misc{hu2021loralowrankadaptationlarge,
      title={LoRA: Low-Rank Adaptation of Large Language Models},
      author={Edward J. Hu and Yelong Shen and Phillip Wallis and Zeyuan Allen-Zhu and Yuanzhi Li and Shean Wang and Lu Wang and Weizhu Chen},
      year={2021},
      eprint={2106.09685},
      archivePrefix={arXiv},
      primaryClass={cs.CL},
      url={https://arxiv.org/abs/2106.09685},
}

@misc{goodfellow2015explainingharnessingadversarialexamples,
      title={Explaining and Harnessing Adversarial Examples},
      author={Ian J. Goodfellow and Jonathon Shlens and Christian Szegedy},
      year={2015},
      eprint={1412.6572},
      archivePrefix={arXiv},
      primaryClass={stat.ML},
      url={https://arxiv.org/abs/1412.6572},
}

@misc{carlini2017evaluatingrobustnessneuralnetworks,
      title={Towards Evaluating the Robustness of Neural Networks},
      author={Nicholas Carlini and David Wagner},
      year={2017},
      eprint={1608.04644},
      archivePrefix={arXiv},
      primaryClass={cs.CR},
      url={https://arxiv.org/abs/1608.04644},
}

\clearpage
\appendix
\section*{Supplementary Material}
\addcontentsline{toc}{section}{Supplementary Material}

\section{Detailed Analysis of Hallucination Causes}
\label{sec:analyse}
We analyze the underlying causes of object hallucination in MLLMs from the perspective of visual features. We use LLaVA v1.5 7B as a representative MLLM in the following analysis.

\subsection{Quantitative Analysis} To quantitatively assess the alignment between visual feature and ground truth text, we compute the cosine similarity between the image embedding $\mathbf{v}$ and the corresponding ground truth text embedding $\mathbf{t}$ of MLLM.
The cosine similarity is calculated as:
\begin{equation}
\text{CosineSim}(\mathbf{v}, \mathbf{t}) = \frac{\mathbf{v} \cdot \mathbf{t}}{|\mathbf{v}| \cdot |\mathbf{t}|},
\end{equation}
Higher similarity indicates better alignment between visual features and textual semantics. By comparing the average similarity across correctly predicted and hallucinated cases, we evaluate how the alignment between visual features and text semantics correlates with hallucination occurrences.

In the analysis, we do not measure similarity between the image and an arbitrary input question. Instead, for each image, we first construct 10 POPE-style questions(Is there a [object] in the image?) based on the ground-truth annotations in COCO, and prompt the model to answer them. If all questions are answered correctly, we label the image as non-hallucinated; otherwise, we label it as hallucinated. Next, we randomly sample 500 correct cases and 500 hallucinated cases from COCO dataset~\cite{lin2015microsoftcococommonobjects}.  For each sampled image, we take its ground-truth caption and compute the cosine similarity between the image embedding and ground-truth text embedding. As shown on the left of Figure 1, correct cases exhibit significantly higher text-image embedding similarity compared to hallucinated cases. The average similarity for correct cases is 0.158, while for hallucinated cases it drops to -0.122. This clear gap suggests that when hallucination occurs, the model's image representation is less aligned with the ground truth text, indicating inaccurate visual perception.

\subsection{Qualitative Analysis}
In qualitative analysis, we adopt Smooth Grad-CAM \cite{zhang2024redundancyrelevanceinformationflow} techniques to visualize the attention maps of MLLMs on non-hallucinated versus hallucinated images.

Given the output logits $\mathbf{z} = [z_1, z_2, \dots, z_n]$, we sum the logits to obtain:
\begin{equation}
z_{\text{answer}} = \sum_{i=1}^{n} z_i
\end{equation}

For the target layer's feature maps $A^k$, we compute the gradients:
\begin{equation}
G^k = \frac{\partial z_{\text{answer}}}{\partial A^k}
\end{equation}

Then, global average pooling derives channel weights:
\begin{equation}
\alpha_k = \frac{1}{Z} \sum_{i,j} G^k_{i,j}
\end{equation}

where $Z$ is the spatial resolution of $A^k$.

The Grad-CAM map is calculated as:
\begin{equation}
M_{\text{Grad-CAM}} = \text{ReLU}\left(\sum_k \alpha_k A^k\right)
\end{equation}

For Smooth Grad-CAM, input image noise $\epsilon_i \sim \mathcal{N}(0, \sigma^2)$ is added, and the Grad-CAM maps are averaged over $N$ samples:
\begin{equation}
M_{\text{Smooth Grad-CAM}} = \frac{1}{N} \sum_{i=1}^{N} M_{\text{Grad-CAM}}^{(i)}
\end{equation}

We further analyze the model's internal attention using Smooth Grad-CAM. The right side of Figure 1 presents attention heatmaps for two non-hallucinated and two hallucinated examples. In non-hallucinated cases, the model focuses accurately on the main object, so that it correctly recognizes existing objects and points out nonexistent objects. In contrast, hallucinated cases show two distinct failure patterns: (1) the model incorrectly focuses on objects visually similar to the target and perceives nonexistent objects (third case); (2) the model focuses on irrelevant regions, neglecting the actual object (fourth case). These patterns reveal that object hallucination often stems from misdirected or insufficient attention to relevant visual features.

\subsection{Summary} Both quantitative and qualitative results show that object hallucination in MLLMs is closely linked to incorrect visual feature perception. Compared to correct cases, hallucinated instances often have lower text-image embedding similarity and misaligned attention distributions. Inspired by these findings, we propose an adversarial fine-tuning framework that explicitly contrasts hallucinated and correct cases to help the model learn more accurate visual representations and mitigate object hallucination.

\subsection{Causal Interventions on the Visual Encoding Stage}
\label{sec:causal}
The quantitative and qualitative analyses above establish a strong \emph{association} between visual feature quality and hallucination. To further verify that this relationship is \emph{causal} rather than merely correlational, we directly intervene on the visual encoding stage and observe the effect on hallucination, using POPE as the testbed and LLaVA-v1.5-7B as the backbone. We perturb visual feature quality in both directions. To degrade it, we (1) add Gaussian noise to the input image, (2) downsample the input to $112\times112$ (to rule out the possibility that the accuracy drop is merely due to out-of-distribution noise), and (3) replace the original CLIP ViT-L/14 with a weaker CLIP ViT-B/16. To strengthen it, we (4) replace the encoder with a stronger SigLIP-SO400M. We report the average accuracy on the three POPE subsets in Table~\ref{tab:causal}.

\begin{table}[t!]
\centering
\resizebox{0.85\columnwidth}{!}{%
\begin{tabular}{lc}
\toprule
\textbf{Intervention on visual encoding} & \textbf{POPE Acc.} \\
\midrule
Base (unchanged) & 0.842 \\
\midrule
\quad (1) Gaussian-noised input & 0.739 \\
\quad (2) Downsampled input ($112\times112$) & 0.751 \\
\quad (3) Weaker encoder (CLIP ViT-B/16) & 0.822 \\
\midrule
\quad (4) Stronger encoder (SigLIP-SO400M) & \textbf{0.864} \\
\bottomrule
\end{tabular}%
}
\caption{Causal interventions on the visual encoding stage (LLaVA-v1.5-7B, average POPE accuracy). POPE accuracy varies monotonically with visual signal quality.}
\label{tab:causal}
\end{table}

POPE accuracy varies monotonically with the quality of the visual signal: each degradation lowers it, whereas the stronger encoder raises it. This causal evidence supports our claim that visual feature quality is a genuine driver of object-existence hallucination, not merely a correlate.

\section{Implementation Details for Baseline Methods}
\label{sec:impl1}
In this section, we also provide the detailed hyperparameter settings used for each baseline. Most hyperparameters are kept consistent with those reported in the original papers. For VCD, we set the noise step to 500, \texttt{cd-alpha} to 1, and \texttt{cd-beta} to 0.1. For OPERA, we use a scale factor of 50.0, set the OPERA threshold to 15, the number of attention candidates to 5, the penalty weights to 1.0, and apply beam search with 5 beams. For Woodpecker, we adopt GroundingDINO~\cite{liu2024groundingdinomarryingdino} as the detector model, setting the box threshold to 0.35 and the text threshold to 0.25. For VTI, we set $\alpha$ to 0.2, $\beta$ to 0.4, the beam search size to 4, and the mask ratio to 0.99.

\section{Implementation Details for AHAF and ACFT}
\label{sec:impl2}
We present the detailed hyperparameters used in AHAF and ACFT. For AHAF, we mainly employed PGD to generate adversarial perturbation and set the number of iterations to 100, the attack budget $\epsilon$ to 16 / 255, and the step size $\alpha$ to 1.
For ACFT, we fine-tuned the target MLLMs using LoRA~\cite{hu2021loralowrankadaptationlarge}.  We kept the text tokenizer frozen and fine-tuned the language model, vision-language connector, and vision encoder. We trained MLLMs on 3,000 samples for 2 epochs with a batch size of 16. The learning rates are set to 2e-4 for the language model, 2e-5 for the vision-language connector, and 1e-5 for the vision encoder. For $\lambda$ in Equation(8), we conduct a hyperparameter search in the range {0.1, 0.2, 0.25, 0.3, 0.4, 0.5, 1.0}, and set $\lambda = 0.25$ based on the empirical results. For $\tau$ in Equation(4), we also conduct a hyperparameter search in the range {0.01, 0.03, 0.05, 0.07, 0.1, 0.2}, and set $\tau = 0.05$ based on the empirical results. All training is conducted on a single NVIDIA H100 GPU.

\section{Experimental Setup and Results for Comparison between OCFT and ACFT}
\label{sec:ocftdetail}
In this section, we present the experimental setup and supplementary results for Section: \textit{Comparison between OCFT and ACFT}.

We randomly sampled 50 images from the COCO dataset~\cite{lin2015microsoftcococommonobjects} as positive images for both methods. For ACFT, we applied PGD attack to generate corresponding negative samples. For OCFT, negative samples were randomly selected from the COCO dataset.

We used LLaVA v1.5~\cite{liu2023llava} to extract embeddings for both positive and negative samples, denoted as $z_{A}^+$, $z_{A}^-$, $z_{O}^+$, and $z_{O}^-$, respectively. Then we selected a target object (i.e., the object intended to induce hallucination) along with several irrelevant non-target objects. For each, we obtained the corresponding anchor text embeddings, denoted as $t^{tar}$ and $t^{non}$. We then computed the cosine similarity between each image embedding and text embedding using cosine similarity $\psi(t, z)$, and get $sim^+ = \psi(t, z^+)$ and $sim^- = \psi(t, z^-)$. The similarity gap $\Delta$ was calculated as $\Delta = |sim^+ - sim^-|$.

For ACFT, the average similarity gap for target object $\Delta_{A}^{tar}$  is 0.738, while the average similarity gap for non-target objects $\Delta_{A}^{non}$ is 0.276, which shows an evident distinction. And the similarity gap of target objects is clearly much higher than non-target ones. In contrast, OCFT yields $\Delta_{O}^{tar} = 0.104$ and $\Delta_{O}^{non} = 0.121$, which are nearly indistinguishable. These results suggest that ACFT can effectively manipulate the model's perception of the target object while exerting minimal influence on non-target objects---an ability that OCFT fails to achieve. The results partially explain why ACFT achieves superior performance compared to OCFT.

\subsection{Counterfactual: Visual-Origin vs.\ Language-Prior Negatives}
\label{sec:counterfactual}
To verify that the gains of ACFT are specifically rooted in the visual-origin diagnosis rather than a generic adversarial-robustness effect, we run a counterfactual experiment that keeps the ACFT pipeline fixed and only swaps the source of the negatives. Using the same 6{,}000 COCO images, we replace AHAF's visual-origin negatives with language-prior negatives: hallucinated captions generated by GPT-5.5. We then fine-tune LLaVA-v1.5-7B under the identical ACFT setup and report the average accuracy on the three POPE subsets. As shown in Table~\ref{tab:counterfactual}, the base model reaches 0.842; the language-prior variant reaches only 0.849, whereas ACFT with AHAF visual negatives attains 0.881. The gain thus depends on the negatives being visual-origin: swapping in language-prior negatives almost eliminates it. This rules out a generic adversarial-robustness effect and confirms that ACFT's gains are rooted in the visual-origin diagnosis, consistent with the OCFT ablation above where unaligned visual negatives also fail.

\begin{table}[t!]
\centering
\resizebox{0.9\columnwidth}{!}{%
\begin{tabular}{lc}
\toprule
\textbf{Negative source (ACFT pipeline fixed)} & \textbf{POPE Acc.} \\
\midrule
Base & 0.842 \\
Language-prior negatives (GPT-5.5 captions) & 0.849 \\
\rowcolor{lightpurple} AHAF visual-origin negatives (ours) & \textbf{0.881} \\
\bottomrule
\end{tabular}%
}
\caption{Counterfactual on the source of negatives (LLaVA-v1.5-7B, average POPE accuracy). The improvement is driven by visual-origin negatives.}
\label{tab:counterfactual}
\end{table}


\section{Details about Adversarial Example Generation in AHAF}

For the hyperparameter settings in AHAF, we select $\epsilon$ from $\{4/255,$ $8/255,$ $16/255,$ $32/255,$ $64/255\}$ and the number of iterations from $\{50, 100, 200, 500, 1000\}$, with the step size fixed at 1. With different hyperparameter settings, we performed PGD attack on 50 randomly selected images. We observe that when $\epsilon$ is too small, the number of iterations required to successfully flip the hallucinated attribute becomes very high, and manual inspection is often required to make sure the attack is successful. Conversely, when $\epsilon$ is too large, the adversarial noise becomes visually apparent, and the generated image significantly deviates from the target. To balance visual quality and attack effectiveness, we set $\epsilon$ to 16/255 and the number of iterations to 100.

Compared to two commonly used adversarial attack methods, FGSM~\cite{goodfellow2015explainingharnessingadversarialexamples} and CW~\cite{carlini2017evaluatingrobustnessneuralnetworks}, PGD offers a favorable trade-off between effectiveness and computational efficiency in the AHAF task. FGSM applies a single-step perturbation, making it less robust and often yielding lower attack success rates in more challenging scenarios. On the other hand, the CW attack involves complex optimization procedures and is computationally expensive. In contrast, PGD is both simpler to implement and more scalable to large-scale datasets and models, while still delivering strong attack performance. Therefore, we adopt PGD as the primary adversarial method in our AHAF framework.

\subsection{Locus of AHAF Perturbations}
\label{sec:locus}
To examine whether AHAF perturbations primarily affect the visual pathway rather than merely exploiting text-side token dynamics, we perform an additional sanity check on 500 successfully flipped contrastive pairs. For each clean image $X$ and its adversarial counterpart $X_{\text{adv}}$, we measure the relative representation change at two stages:
\begin{equation}
\varepsilon_{\text{vis}} = \frac{\|h_{\text{vis}}^{\text{adv}} - h_{\text{vis}}^{\text{clean}}\|_2}{\|h_{\text{vis}}^{\text{clean}}\|_2}, \qquad
\varepsilon_{\text{LLM}} = \frac{\|h_{\text{LLM}}^{\text{adv}} - h_{\text{LLM}}^{\text{clean}}\|_2}{\|h_{\text{LLM}}^{\text{clean}}\|_2},
\end{equation}
where $h_{\text{vis}}$ denotes the projected visual token embeddings and $h_{\text{LLM}}$ the LLM text-token hidden states measured at the deep (post-attention) layers, so that the text tokens have already integrated visual information. Averaged over all flipped samples, we obtain $\varepsilon_{\text{vis}} = 0.057$ and $\varepsilon_{\text{LLM}} = 0.026$. The representation change at the visual-token level is over $2\times$ larger than that observed in the text-token hidden states. This result suggests that AHAF perturbations are concentrated in the visual representation rather than the language model's text-token processing, which supports our claim.

\section{Computational Cost Analysis}
\label{sec:cost}
In this section, we provide detailed analysis for both AHAF and ACFT stages.

In the AHAF stage, we construct a training set with 3,000 contrastive image samples and 3,000 normal samples. For the 3,000 contrastive samples, we run PGD on each image with 100 iterations. On a single NVIDIA A100, the average optimization time per image is about 15s, so constructing all 3,000 adversarial images takes roughly 12 GPU-hours.

In the ACFT stage, the contrastive loss only introduces only a lightweight additional computation overhead on top of standard SFT: we reuse the last-layer image and text representations to compute the adversarial contrastive loss, without extra forward passes through the backbone. In practice, for LLaVA-v1.5-7B with batch size 16 on a single A100, plain SFT training on our 6k-sample set takes 30 min 57s, while ACFT takes 32 min 21s---an overhead of about 90s, which we consider negligible. For comparison, when we run CHiP-DPO~\cite{fu2025chip} using the official training script on the same backbone, training requires 3 h 21 min 45 s on 4 A100 GPUs, i.e., more than 13 GPU-hours, on a dataset of comparable size. We also apply vanilla DPO to fine-tune LLaVA v1.5-7B on 6000 COCO images. The DPO training process takes 1 h 33 min 27 s on 4$\times$A100 GPUs (about 6 GPU-hours), whereas ACFT only requires about 0.5 GPU-hours.

\section{Comparison with Training-based Baselines}
\label{sec:trainbaseline}
Beyond the inference-time methods compared in the main text, we further compare ACFT against representative training-based approaches under a strictly controlled protocol: standard SFT, standard adversarial training, standard DPO, and CHiP-DPO~\cite{fu2025chip}. All methods use the same backbone (LLaVA-v1.5-7B), the same training data (6{,}000 samples from COCO with generated negatives), and the same training budget. We report the average accuracy on the three POPE subsets in Table~\ref{tab:trainbaseline}.

\begin{table}[t!]
\centering
\resizebox{0.85\columnwidth}{!}{%
\begin{tabular}{lc}
\toprule
\textbf{Method (same backbone/data/budget)} & \textbf{POPE Acc.} \\
\midrule
Base & 0.842 \\
Standard SFT & 0.851 \\
Standard adversarial training & 0.848 \\
Standard DPO & 0.849 \\
CHiP-DPO & 0.856 \\
\rowcolor{lightpurple} ACFT (ours) & \textbf{0.881} \\
\bottomrule
\end{tabular}%
}
\caption{Comparison with training-based baselines under identical backbone (LLaVA-v1.5-7B), data (6K COCO with generated negatives), and budget. Average POPE accuracy.}
\label{tab:trainbaseline}
\end{table}
Under identical backbone, data, and budget, ACFT still delivers the best performance. A likely reason is that only ACFT directly targets the root cause of image--text misalignment: AHAF's aligned pairs isolate the hallucination-triggering visual features, and the contrastive loss explicitly corrects this cross-modal gap. In contrast, SFT and DPO supervise only at the output or preference level, while adversarial training targets perturbation robustness rather than cross-modal alignment; none directly closes the image--text gap.

\section{Generalization to Additional Architecture}
\label{sec:newmodel}
To further demonstrate that ACFT is architecture-agnostic and generalizes well beyond the models evaluated in the main experiments, we applied ACFT to InternVL-3.5-4B~\cite{wang2025internvl35advancingopensourcemultimodal}, a recent MLLM with a distinct architecture from LLaVA, MiniGPT-4, and Qwen2.5-VL. Results are shown in Table~\ref{tab:internvl}. ACFT delivers consistent gains on POPE across all three subsets, improving accuracy by 1.3\%, 1.3\%, and 0.4\%, respectively. On MME Existence, both the original model and ACFT achieve perfect accuracy of 1.000, reflecting that this task is relatively easy for modern VLMs. On MME Whole, ACFT also brings a marginal improvement. These results confirm that ACFT transfers effectively to diverse model architectures without any architecture-specific modifications, supporting its broad applicability.

\begin{table}[t!]
\centering
\resizebox{\columnwidth}{!}{%
\begin{tabular}{ccccccc}
\toprule
\textbf{Bench.} & \textbf{Subset} & \textbf{Method} & \textbf{ACC} & \textbf{Pre.} & \textbf{Rec.} & \textbf{F1} \\
\midrule
\multirow{6}{*}{POPE}
  & \multirow{2}{*}{\textit{Adv.}}
      & original & 0.863 & 0.835 & \textbf{0.905} & 0.869 \\
  &   & \cellcolor{lightpurple}ACFT & \cellcolor{lightpurple}\textbf{0.876} & \cellcolor{lightpurple}\textbf{0.864} & \cellcolor{lightpurple}0.893 & \cellcolor{lightpurple}\textbf{0.878} \\
\cmidrule(lr){2-7}
  & \multirow{2}{*}{\textit{Pop.}}
      & original & 0.899 & 0.894 & \textbf{0.905} & 0.899 \\
  &   & \cellcolor{lightpurple}ACFT & \cellcolor{lightpurple}\textbf{0.912} & \cellcolor{lightpurple}\textbf{0.927} & \cellcolor{lightpurple}0.895 & \cellcolor{lightpurple}\textbf{0.911} \\
\cmidrule(lr){2-7}
  & \multirow{2}{*}{\textit{Rand.}}
      & original & 0.933 & \textbf{0.969} & 0.845 & 0.930 \\
  &   & \cellcolor{lightpurple}ACFT & \cellcolor{lightpurple}\textbf{0.937} & \cellcolor{lightpurple}0.966 & \cellcolor{lightpurple}\textbf{0.905} & \cellcolor{lightpurple}\textbf{0.935} \\
\midrule
\multirow{4}{*}{MME}
  & \multirow{2}{*}{\textit{Exist.}}
      & original & \textbf{1.000} & \textbf{1.000} & \textbf{1.000} & \textbf{1.000} \\
  &   & \cellcolor{lightpurple}ACFT & \cellcolor{lightpurple}\textbf{1.000} & \cellcolor{lightpurple}\textbf{1.000} & \cellcolor{lightpurple}\textbf{1.000} & \cellcolor{lightpurple}\textbf{1.000} \\
\cmidrule(lr){2-7}
  & \multirow{2}{*}{\textit{Whole}}
      & original & 0.859 & \textbf{0.913} & 0.795 & 0.850 \\
  &   & \cellcolor{lightpurple}ACFT & \cellcolor{lightpurple}\textbf{0.862} & \cellcolor{lightpurple}0.894 & \cellcolor{lightpurple}\textbf{0.821} & \cellcolor{lightpurple}\textbf{0.856} \\
\bottomrule
\end{tabular}%
}
\caption{InternVL-3.5-4B: original vs. ACFT on POPE and MME.}
\label{tab:internvl}
\end{table}

\section{Generalization to Counting Hallucination}
\label{sec:counting}
To demonstrate that ACFT transfers well to short-answer tasks other than yes/no object-existence questions, we further evaluate it on counting hallucination, where the model misperceives the number of objects in an image. We randomly sample 1{,}000 examples from TallyQA as the test set and use LLaVA-v1.5-7B as the base model. The base model achieves an accuracy of 0.673 on this test set, whereas our ACFT model achieves 0.705, a gain of over 3 points, even though ACFT is never specifically trained on counting tasks. This indicates that the gains reflect improved visual grounding rather than memorization of the training format, and that ACFT generalizes beyond the yes/no format.

\end{document}